\documentclass[review,5p,times]{elsarticle}
\usepackage[T1]{fontenc}
\usepackage[utf8]{inputenc}
\usepackage{adjustbox}
\usepackage{makecell}
\usepackage{amssymb}
\usepackage{amsmath}
\usepackage{lineno}
\usepackage{array}
\usepackage{rotating}
\usepackage{float}
\usepackage[ruled,vlined,linesnumbered]{algorithm2e}
\usepackage{subcaption}
\usepackage{tablefootnote}
\usepackage{url}
\usepackage{footmisc}
\usepackage[colorlinks = true, 
   linkcolor = blue,
   urlcolor = blue,
   citecolor = blue,
   anchorcolor = blue]{hyperref}

\usepackage{longtable}
\usepackage{ltablex}
\keepXColumns

\usepackage{etoolbox}
\usepackage{pdflscape}
\usepackage{pifont}
\usepackage{tcolorbox}
\usepackage{tikz}
\usepackage{booktabs}
\usepackage{colortbl}
\usepackage{enumitem}

\usepackage{amsthm}
\newtheorem{lemma}{Lemma}
\newtheorem{theorem}{Theorem}
\SetNlSkip{0.8em}   

\begin{document}

\begin{frontmatter}

\title{UAMDP: Uncertainty-Aware Markov Decision Process for Risk-Constrained Reinforcement Learning from Probabilistic Forecasts}

\author[1]{Michal Koren}
\author[1]{Or Peretz}
\author[2]{Tai Dinh\corref{mycorrespondingauthor}}
\cortext[mycorrespondingauthor]{Corresponding Author}
\ead{t\_dinh@kcg.ac.jp}
\author[3]{Philip S. Yu}

\address[1]{Shenkar-Engineering. Design. Art., Ramat Gan, 252626, Israel}
\address[2]{The Kyoto College of Graduate Studies for Informatics, Sakyo, Kyoto City, Kyoto, 606-8225, Japan}
\address[3]{Department of Computer Science, University of Illinois at Chicago, Chicago, 60607-7045, USA}

\begin{abstract}
Sequential decisions in volatile, high-stakes settings require more than maximizing expected return; they require principled uncertainty management. This paper presents the Uncertainty-Aware Markov Decision Process (UAMDP), a unified framework that couples Bayesian forecasting, posterior-sampling reinforcement learning, and planning under a conditional value-at-risk (CVaR) constraint. In a closed loop, the agent updates its beliefs over latent dynamics, samples plausible futures via Thompson sampling, and optimizes policies subject to preset risk tolerances. We establish regret bounds that converge to the Bayes-optimal benchmark under standard regularity conditions. We evaluate UAMDP in two domains including high-frequency equity trading and retail inventory control, both marked by structural uncertainty and economic volatility. Relative to strong deep learning baselines, UAMDP improves long-horizon forecasting accuracy (RMSE decreases by up to 25\% and sMAPE by 32\%), and these gains translate into economic performance: the trading Sharpe ratio rises from 1.54 to 1.74 while maximum drawdown is roughly halved. These results show that integrating calibrated probabilistic modeling, exploration aligned with posterior uncertainty, and risk-aware control yields a robust, generalizable approach to safer and more profitable sequential decision-making.
\end{abstract}


\end{frontmatter}


\section{Introduction} \label{introduction}
Robust decision-making under uncertainty is critical in dynamic, high-stakes domains such as finance, healthcare, and energy systems, where even marginally suboptimal actions can incur substantial economic or human costs. These environments are not only volatile but also structurally complex, often exhibiting latent dynamics that challenge both predictive accuracy and policy reliability. As such, effective decision frameworks must go beyond deterministic predictions and pointwise optimization to integrate mechanisms for uncertainty quantification and risk-sensitive control \cite{hsieh2025risk}.

Despite significant advances, most traditional reinforcement learning (RL) approaches to forecasting and decision-making remain focused on maximizing expected returns. This narrow objective often yields policies that are overconfident in their assumptions and brittle in distributional shifts or unanticipated conditions. Meanwhile, state-of-the-art forecasting models frequently produce deterministic or under-calibrated outputs, diminishing their value for downstream decisions that depend on uncertainty awareness \cite{lim2021temporal}. These parallel deficiencies highlight a fundamental gap: the lack of a cohesive, uncertainty-aware loop that connects probabilistic modeling with sequential control.

Recent progress in probabilistic forecasting, particularly using Gaussian processes (GPs) and Temporal Fusion Transformers (TFTs), has demonstrated the feasibility of generating calibrated predictive intervals, offering a more nuanced view of future states \cite{lim2021temporal}.

Yet in practice, these models are rarely integrated into the control layer of forecasting software. Instead, they are treated as peripheral inputs, decoupled from the agent's learning and planning mechanisms. In parallel, deep RL has demonstrated remarkable performance on complex tasks but often relies on heuristic exploration and disregards uncertainty estimates altogether \cite{sutton1998reinforcement}. Consequently, learned policies may perform well under familiar conditions but collapse when confronted with unfamiliar inputs, precisely where uncertainty-aware reasoning is most needed \cite{abdar2021review}.

Moreover, mainstream RL formulations frequently assume fully observable and discretized environments. This abstraction neglects the partial observability and continuous dynamics that characterize many real-world systems, from patient health trajectories to financial markets \cite{ladosz2022exploration}. Even within methods that nominally address uncertainty, the treatment is often superficial, relying on simplistic approximations or failing to propagate uncertainty through the decision pipeline. This methodological gap severely limits the reliability of RL-based systems in environments where uncertainty is inherent and safety, robustness, and risk awareness are not optional but essential \cite{komorowski2018artificial}.

To bridge these gaps, we propose the Uncertainty-Aware Markov Decision Process (UAMDP), a novel framework that unifies probabilistic forecasting and RL in a closed-loop, Bayes-adaptive MDP formulation. The UAMDP maintains a posterior belief over latent environment parameters, which is continuously updated using forecasts from state-of-the-art probabilistic models-GPs for moderate-scale problems and Monte Carlo dropout transformers for high-dimensional time series. These beliefs drive posterior-sampling RL, enabling the agent to simulate plausible futures and optimize policies that are both exploratory and risk-aware \cite{osband2013more}. The primary contributions of this work are summarized below:

\begin{itemize}
\item The UAMDP integrates Bayesian forecasting directly into the RL process, enabling real-time belief updates and adaptive control in uncertain environments.
\item Through sampling from the agent's posterior belief, the framework supports principled exploration that prioritizes regions of high epistemic uncertainty. This mechanism achieves near-optimal regret when the planning horizon matches the episode length and each parameter setting is sampled at least once.
\item Incorporating CVaR-constrained planning allows the agent to mitigate downside risk and enhances robustness to rare but high-impact events.
\item To maintain tractability in high-dimensional settings, the framework employs sparse GPs and TFTs, thereby extending its applicability to large-scale datasets without sacrificing calibration fidelity.
\end{itemize}

The remainder of this paper is organized as follows. Section~\ref{related_work} surveys the state of the art in probabilistic time-series forecasting, uncertainty-aware RL, and integrated forecast-control architectures. Section~3 formalizes the UAMDP as a Bayes-adaptive MDP and presents the Bayesian forecasting module, posterior-sampling loop, and risk-sensitive extensions. Section~4 describes the empirical design, including datasets, baselines, and evaluation metrics. Section~5 reports sample empirical quantitative and qualitative results, emphasizing improvements in calibration, decision performance, and robustness with respect to input perturbations. Section~6 concludes with a discussion of practical implications, limitations, and future research directions.

\section{Related Work} \label{related_work}
The development of the UAMDP builds upon three interconnected research threads: probabilistic time-series forecasting, uncertainty-aware RL, and the integration of predictive distributions into sequential decision-making. This section synthesizes key advances in each area to establish the conceptual scaffolding for our framework. We begin by discussing the evolution of probabilistic forecasting models, including GPs, deep generative models, and transformer-based architectures. We then examine RL approaches that explicitly model epistemic and aleatoric uncertainty. Finally, we explore recent efforts to bridge predictive modeling and control, highlighting why a fully Bayesian, closed-loop integration remains an open and necessary challenge.

\subsection{Probabilistic time-series forecasting}
Bayesian approaches have long emphasized the role of predictive distributions in sequential decision-making. GP models have advanced non-parametric forecasting by offering analytically tractable uncertainty quantification, establishing themselves as foundational tools in high-volatility domains such as finance and critical care \cite{liu2021incremental}. Recent work on dynamic feature-fusion architectures further extends these ideas to large-scale time series with rich covariates, for example the DFF-Net model of \cite{xiao2025dff}, which learns to adaptively combine multiple feature channels over time.

The emergence of deep learning has significantly extended the capabilities of probabilistic forecasting, enabling accurate modeling across large-scale and multi-series datasets. Frameworks such as DeepAR, which employ autoregressive recurrent neural networks (RNNs) to directly estimate likelihood parameters, have achieved notable accuracy gains, reportedly reducing forecast error by up to 15\% in domains such as retail and energy \cite{salinas2020deepar,koren2022inventory,koren2024inventory}. Further advances, including TFTs with Monte Carlo dropout \cite{lim2021temporal} and multi-horizon quantile RNNs \cite{gasthaus2019probabilistic}, have enriched forecast calibration, while recent regression-based models address interval misestimation through post-hoc correction. Hybrid approaches that combine data-driven models with structured prior knowledge offer another promising direction for robust probabilistic forecasting \cite{chen2024towards}.

Latent-state forecasting frameworks have also shown considerable promise by integrating deep neural architectures with Bayesian filtering. Variational RNNs \cite{chung2015gated}, deep Markov models \cite{fraccaro2016sequential}, and deep state-space models \cite{rangapuram2018deep} capture latent regime transitions that often evade purely deterministic models. Complementary approaches, such as Bayesian ensembling and model stacking, alongside classical methods such as quantile regression and interval prediction \cite{kabir2018partial}, further reduce model risk. Related work in approximate reasoning has studied imprecise and non-stationary uncertainty models for sequential predictions \cite{persiau2022dis}, emphasizing the importance of credible interval estimates. Collectively, these contributions underscore the necessity of generating credible predictive intervals, not only to improve forecasting accuracy but to enable safe and informed downstream decision-making under uncertainty.

\subsection{Reinforcement learning under uncertainty}
RL research has long addressed the inherent challenges posed by stochastic rewards and model uncertainty. Probably Approximately Correct (PAC) MDP algorithms, such as R-MAX  \cite{brafman2002r} and UCRL \cite{auer2008near}, offer finite-sample performance guarantees by leveraging optimism in the face of uncertainty. More recent advances in distributional RL model the full return distribution rather than expected values, enabling richer value representations \cite{dabney2018distributional}. In parallel, risk-sensitive RL has introduced tail-aware optimization objectives such as CVaR, replacing the conventional focus on expected returns with measures of downside exposure \cite{tamar2015optimizing}. Complementary lines of work in approximate reasoning and decision theory study risk-sensitive Markov decision processes with budget constraints \cite{moreira2024efficient} and three-way decision models that explicitly encode different attitudes toward risk and uncertainty \cite{xin2025interval,li2025improved,suo2024review}.

Domain-specific studies have further demonstrated the practical importance of uncertainty quantification. In autonomous driving, posterior variance estimates are used to improve collision-avoidance strategies, particularly in high-risk scenarios such as intersection navigation \cite{hoel2020reinforcement}. In finance, uncertainty-aware policies have incorporated variational autoencoder-based epistemic penalties to enhance portfolio resilience, complemented by hierarchical controllers and normalizing-flow transition models capable of adapting to market shocks \cite{enkhsaikhan2024uncertainty,koren2023forecasting}. Despite these advances, recent reviews underscore persistent barriers to industrial deployment\textemdash{}most notably, limited robustness under distributional shift and the absence of standardized evaluation protocols \cite{hambly2023recent}.

Bayesian RL formalizes exploration--exploitation trade-offs via probabilistic inference \cite{ghavamzadeh2015bayesian}. Posterior-sampling RL, achieving near-optimal regret, has gained attention due to its effective policy optimization \cite{osband2013more}. Extensions such as Bayesian Q-learning \cite{dearden1998bayesian}, variance-adaptive Thompson sampling \cite{thompson1933likelihood}, and meta-contextual bandits broaden applicability \cite{koren2024automated}. The probabilistic inference for learning control (PILCO) framework notably demonstrates data-efficient Bayesian policy learning by leveraging GP moment propagation \cite{deisenroth2011pilco}, with Monte Carlo and dropout-based variants addressing partial observability and global stability.

Recent contributions to Bayesian RL include Bayesian exploration networks (BEN), which enable model-free learning of value distributions under uncertainty, and Bayesian residual policy optimization, which incorporates expert priors to refine action selection \cite{lee2021bayesian}. These innovations represent substantial progress toward Bayes-optimal decision-making, particularly in high-stakes settings where uncertainty plays a central role. In parallel, learning human-interpretable decision catalogues and scoring systems for situated decision making has attracted renewed interest \cite{heid2024learning}, reflecting a broader shift toward transparent policies. Notably, the integration of probabilistic forecasting with RL has demonstrated promising results across a range of application domains.

In finance, DeepAR-based predictive distributions have been employed to guide portfolio optimization, yielding substantial improvements in risk-adjusted returns \cite{wen2017multi}. In healthcare, Bayesian methods have enhanced intensive care unit risk scoring and sepsis treatment through calibrated probabilistic forecasting \cite{komorowski2018artificial} and have been further applied to personalized insulin dosing using CVaR-constrained optimization \cite{lu2023structured}. In autonomous systems, forecast-informed RL has improved decision robustness under sensor noise and uncertainty, reducing accident risk and enhancing navigational performance \cite{hoel2020reinforcement}. Similarly, energy management systems have benefited from multiscale forecasting integrated with RL, enabling more effective demand response strategies and reducing peak load stress \cite{yang2025derl}.

\subsection{Integrated forecasting and control}
Classical model-predictive control (MPC) has long incorporated forecast information to guide horizon-based decision-making \cite{rawlings2020model}. Contemporary extensions, including stochastic and chance-constrained MPC, explicitly model uncertainty to enhance control robustness \cite{kouvaritakis2015model}. Deep RL approaches have further advanced this integration. Methods such as probabilistic ensembles with trajectory sampling (PETS) combine probabilistic ensemble dynamics models with trajectory-sampling MPC, achieving competitive performance with substantially improved sample efficiency \cite{chua2018deep}. 

More recent frameworks, such as model predictive path integral control \cite{williams2017model} and Dreamer-V2 \cite{hafner2019dream}, expand this paradigm by leveraging learned dynamics models for real-time control. Bayesian adaptive deep RL methods integrate neural architectures with Thompson sampling, providing strong empirical performance alongside theoretical regret guarantees. These developments align with broader efforts in approximate reasoning and decision theory to design robust decision rules under uncertainty \cite{suo2024review,heid2024learning}. Collectively, these approaches underscore the value of explicitly modeling uncertainty to achieve efficient, high-performance, and robust decision making.

Three key trends emerge from the literature we reviewed. First, probabilistic forecasting has progressed substantially in recent years, yielding robust and calibrated multi-step predictive distributions. Second, RL increasingly incorporates explicit uncertainty quantification, influencing both exploration strategies and safety guarantees. Third, integrated forecasting--control frameworks consistently demonstrate improvements in sample efficiency and risk awareness. Nonetheless, important gaps remain, particularly in scalable belief tracking under partial observability and irregular sampling, unified treatment of aleatoric and epistemic uncertainty, and the availability of standardized benchmarks evaluating both decision quality and forecast calibration. The UAMDP tackles these challenges by integrating Bayesian forecasting with posterior-sampling RL, yielding a framework for sequential decision-making under uncertainty that is both theoretically sound and practically effective.

\section{Uncertainty-aware Markov decision process}

\begin{table}[!h]
\centering
\caption{Table of notations.}
\label{tab:notation}
\begin{tabular}{@{}l p{0.68\linewidth}@{}}
\toprule
\textbf{Symbol} & \textbf{Meaning} \\
\midrule
$\mathcal{S}, \mathcal{A}$ & State and action spaces \\
$x_t \in \mathbb{R}^d$ & Exogenous time-series observation at step $t$ \\
$s_t = (x_{1:t}, a_{0:t-1})$ & Augmented observable history (``state'') \\
$\theta \in \Theta$ & Latent environment parameters \\
$b_t(\theta)$ & Posterior belief over $\theta$ at time $t$ \\
$r_t = r(s_t, a_t)$ & Reward after taking $a_t$ in $s_t$ \\
$\gamma \in (0,1)$ & Discount factor \\
$T, H$ & Global horizon $T$ and episodic horizon $H$ \\
$p_{\phi}(x_{t+1}\mid s_t, a_t, \theta)$ & Probabilistic forecaster (GP or TFT) \\
$\pi_t(\cdot)$ & Policy conditioned on $(s_t, b_t)$ \\
$Z_{\psi}$ & Distributional critic for return random variable $Z$ \\
\bottomrule
\end{tabular}
\end{table}

This section formalizes the proposed UAMDP framework as a Bayes-adaptive MDP, providing a principled foundation for decision-making under uncertainty (see Table~\ref{tab:notation} for definitions of the principal notation used in this study). At its core, the UAMDP integrates probabilistic time-series forecasting with Bayesian RL through a posterior-sampling control loop. This integration enables calibrated uncertainty quantification, promotes sample-efficient exploration, and enforces explicit control over downside risk. 

The framework consists of three core components: (1) a Bayesian forecasting module that continuously updates the agent's belief distribution over latent environment parameters; (2) an RL agent that employs posterior-sampling strategies for adaptive policy generation; and (3) a risk-sensitive planning mechanism that incorporates CVaR constraints to manage downside exposure during decision execution. First, we present the procedural design of the UAMDP, detailing the interaction among forecasting, sampling, and planning. We then provide a formal proof of correctness and derive additive error bounds under bounded approximation assumptions.

We write $\theta$ to index latent environment parameters governing the transition kernel
$P_{\theta}(s' \mid s,a)$. The probabilistic forecaster is parameterized by weights $\phi$,
and the policy and distributional critic are parameterized by weights $\psi$; during online
interaction, $\phi$ and $\psi$ are treated as fixed and the only online model update acts on
the belief $b_t(\theta)$. Since the augmented state $s_t=(x_{1:t},a_{0:t-1})$ contains the
observable history, all dependence on the current observation is captured through $s_t$.
Concretely, in UAMDP the kernel $P_{\theta}$ is induced by the forecaster: given $(s_t,a_t,\theta)$,
we draw $x_{t+1}\sim p_{\phi}(\cdot\mid s_t,a_t,\theta)$ and set
$s_{t+1}=\tau(s_t,a_t,x_{t+1}) := (x_{1:t+1},a_{0:t})$.

\subsection{Procedure flow}
Let $p_{\phi}(x_{t+1}\mid s_t,a_t,\theta)$ denote a probabilistic one-step forecaster for the next observation
$x_{t+1}\in\mathbb{R}^d$. For moderate observation dimension $d$, we instantiate $p_{\phi}$ as a GP; when $d$ is large,
we employ a TFT equipped with Monte-Carlo dropout so that every predictive call returns a full predictive distribution
rather than a point estimate. The forecaster thus supplies the likelihood term required for Bayesian inference over the
latent environment parameter $\theta$.

Specifically, for moderate observation dimension $d$, we instantiate $p_{\phi}(x_{t+1}\mid s_t,a_t,\theta)$ via GP
regression as follows. Recall that $\theta\in\Theta$ denotes latent environment parameters (fixed within an episode) that govern the induced transition kernel $P_{\theta}(s'\mid s,a)$ defined above, while $\phi$ denotes forecaster parameters obtained by offline pre-training and then held fixed during RL training. To define the GP, let $u_t := f(s_t,a_t)\in\mathbb{R}^p$
be a fixed feature map of the observable history and action, and form an augmented GP input
$z_t := \mathrm{concat}(u_t,\eta(\theta))\in\mathbb{R}^{p+p_\theta}$, where $\eta(\theta)$ is an encoding of $\theta$
(e.g., identity for continuous $\theta$, or one-hot/embedding for discrete $\theta$). For each output coordinate
$j\in\{1,\dots,d\}$, we place a GP prior on a latent function
$g_j(\cdot)\sim\mathcal{GP}\!\big(m_{\phi,j}(\cdot),k_{\phi,j}(\cdot,\cdot)\big)$ and assume the observation model
\[
x_{t+1}^{(j)} = g_j(z_t) + \varepsilon_t^{(j)}, \qquad \varepsilon_t^{(j)}\sim \mathcal{N}(0,\sigma_{\phi,j}^2).
\]
After offline training (absorbed into $\phi$), the resulting one-step predictive distribution used in UAMDP is
\[
p_{\phi}(x_{t+1}\mid s_t,a_t,\theta)
= \prod_{j=1}^d \mathcal{N}\!\big(x_{t+1}^{(j)};\mu_{\phi,j}(z_t),\,\Sigma_{\phi,j}(z_t)\big),
\]
where $\mu_{\phi,j}(z_t)$ and $\Sigma_{\phi,j}(z_t)$ are the standard GP posterior mean and variance (with fixed
hyperparameters absorbed into $\phi$); the product form corresponds to independent-output GPs for tractability. This
density provides the likelihood term in the Bayes update of $b_t(\theta)$. Here $\phi$ absorbs the GP specification
learned offline, including the mean and kernel hyperparameters (e.g., length-scales and output variance) and the
observation-noise variance $\sigma_{\phi,j}^2$, as well as any fixed sparsification choices (e.g., inducing inputs, if
used). The functions $\mu_{\phi,j}(z_t)$ and $\Sigma_{\phi,j}(z_t)$ are obtained from the standard GP posterior
predictive equations conditioned on the offline training data (and, if applicable, the inducing representation), so
that $p_{\phi}(\cdot\mid s_t,a_t,\theta)$ is fully specified at test time and the forecaster parameters $\phi$ are not
updated online. Given the history-state definition $s_t=(x_{1:t},a_{0:t-1})$, the induced state-transition kernel is
obtained by first sampling $x_{t+1}\sim p_{\phi}(\cdot\mid s_t,a_t,\theta)$ and then setting
$s_{t+1}=\tau(s_t,a_t,x_{t+1})=(x_{1:t+1},a_{0:t})$, which defines $P_{\theta}(s_{t+1}\mid s_t,a_t)$.

In practice, this choice reflects a trade-off between calibration and computational cost rather than a hard
requirement of UAMDP. Following the common use of Gaussian processes in moderate-scale settings, we instantiate
$p_{\phi}$ as a GP only when both the observation dimension and the available sample size are modest, so that training
and prediction remain tractable. For high-dimensional or large multi-series problems, we instead employ TFT with
Monte-Carlo dropout, as discussed in the empirical section. In that regime, a GP becomes less efficient and more
sensitive to irrelevant covariates, whereas TFT can learn latent representations and capture long-range dependencies
across many input channels. Monte-Carlo dropout at inference time then induces a predictive distribution over
$x_{t+1}$ rather than a single point forecast, and this distribution is plugged into the Bayesian update as the
likelihood term. Importantly, the surrounding UAMDP loop including belief updating, Thompson sampling, and planning
remains unchanged; only the underlying probabilistic forecaster $p_{\phi}$ is swapped.

Initially, the procedure instantiates its three core components. It sets a prior belief $b_{0}(\theta)$ over the latent
environment parameter; selects the forecaster $p_{\phi}(x_{t+1}\mid s_t,a_t,\theta)$, implemented as a GP when the
observation dimension $d$ is moderate or as a TFT with Monte-Carlo dropout when $d$ is high; and loads the
policy/critic weights $\psi$. These elements jointly define the agent's initial information state.

\begin{figure*}
  \centering
  \includegraphics[width=\linewidth]{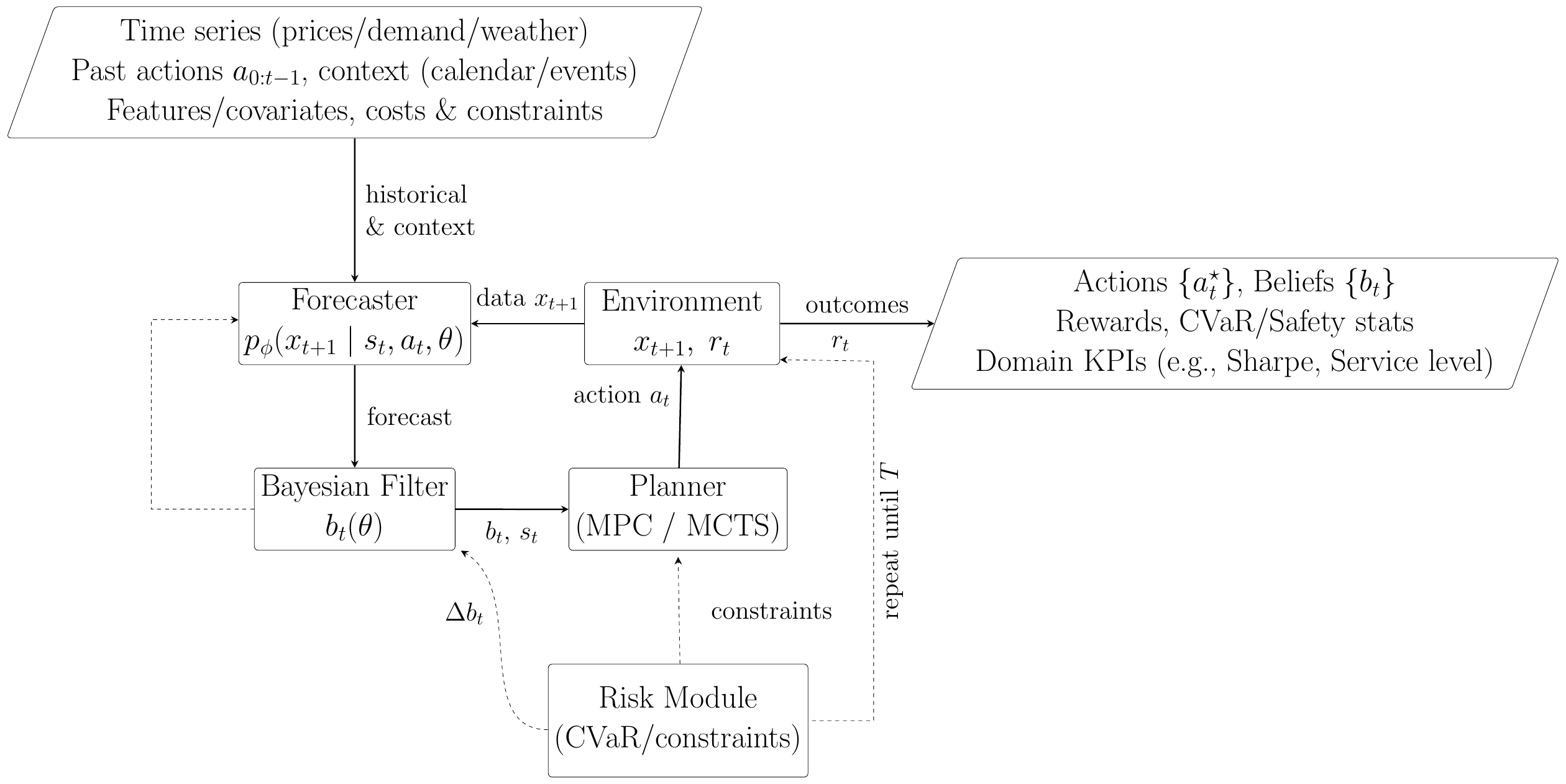}
  \caption{Flow of the proposed framework.}
  \label{fig:work_flow}
\end{figure*}

Next, at the start of each episode, the agent performs Thompson exploration by sampling a concrete model
$\theta_k \sim b(\theta)$, where $b(\theta)$ denotes the current belief (initialized as $b_0(\theta)$). This random draw
aligns the forthcoming policy with the agent's current epistemic uncertainty, thereby encouraging information-seeking
behavior when the belief is diffuse. At each decision step, planning uses a finite lookahead horizon $H$. Conditioned
on the sampled $\theta_k$ and the current hyper-state $(b_t,s_t)$, the agent applies a planner to maximize the Bayesian
action value; in implementation, $\theta_k$ provides a Monte-Carlo approximation for evaluating the belief expectation
inside the planner (e.g., via rollouts under $P_{\theta_k}$). In our empirical study (Section~4), this planner is
instantiated as a bounded-depth Monte Carlo tree search (MCTS); the model-predictive control (MPC) and
differentiable-rollout backends shown in Fig.~\ref{fig:work_flow} illustrate that the UAMDP framework is compatible with
alternative trajectory-optimization controllers, but they are not used for the experiments reported here. The Bayesian
action value is given by
\[
Q(b_t,s_t,a)
= \mathbb{E}_{\theta\sim b_t}\!\left[r(s_t,a)+\gamma\,V^{\pi}\!\big(b_{t+1}^{(\theta)},s_{t+1}\big)\right],
\]
where $V^{\pi}$ denotes the value induced by the rollout/evaluation policy used by the planner (e.g., the default policy
in MCTS).

\begin{algorithm}[!htb]
\caption{UAMDP($b_{0},H,T,F$): online control with Bayesian belief update.}
\label{alg:uamdp}
$b \leftarrow b_{0}$\;
Load forecaster $(F)$ weights $\phi$ and policy/critic weights $\psi$\;
$n \leftarrow 0$\tcp*{global interaction step}
\While{$n < T$}{
  Sample $\theta_k \sim b$\;
  \For{$t \gets 0$ \KwTo $H-1$}{
    \If{$n \ge T$}{\textbf{break}}
    Estimate $Q(b_t,s_t,a)$ under $\theta_k$\;
    \If{risk control is enabled}{
      Replace $Q$ by $\mathrm{CVaR}_{\alpha}[Z_{\psi}(b_t,s_t,a)]$\;
      Enforce $\Pr(x_{t+h}\in S_{\mathrm{safe}})\ge 1-\delta$\;
    }
    $a_t^{\star} \gets \arg\max_{a}\, Q(b_t,s_t,a)$\;
    Execute $a_t^{\star}$ and observe $(x_{t+1},r_t)$\;
    $b_{t+1}(\theta)\propto p_{\phi}(x_{t+1}\mid s_t,a_t^{\star},\theta)\,b_t(\theta)$\;
    $n \leftarrow n+1$\;
  }
  $b \gets b_{t+1}$\;
}
\end{algorithm}

The first action on the resulting optimal sequence, $a_t^{\star}$, is executed while the remainder of the plan is
retained as a receding-horizon estimate. Immediately afterwards, the environment returns the new observation $x_{t+1}$
and reward $r_t$. The belief is refined according to Bayes' rule,
\[
b_{t+1}(\theta)\propto p_{\phi}(x_{t+1}\mid s_t,a_t^{\star},\theta)\,b_t(\theta),
\]
such that parameter values explaining the data gain posterior mass, and future decisions become correspondingly sharper.

Let $(\Theta,\mathcal{T})$ be the measurable space of latent environment parameters and let $\lambda$ be a $\sigma$-finite
reference measure on $\Theta$ (Lebesgue for continuous $\Theta$, counting measure for discrete $\Theta$).
We represent beliefs by densities w.r.t.~$\lambda$ and define the belief space
\[
\mathcal{B}
:= \left\{ b:\Theta\to\mathbb{R}_{\ge 0}\ \middle|\ \int_{\Theta} b(\theta)\,\lambda(d\theta)=1 \right\}.
\]
Equivalently, each $b\in\mathcal{B}$ induces a probability measure $b(d\theta)=b(\theta)\lambda(d\theta)$.
Throughout, $p_\phi(x\mid s,a,\theta)$ denotes the forecaster's predictive density and is assumed measurable in $\theta$
and integrable so that the normalizing constant below is finite and non-zero.
For $(b,s)\in\mathcal{B}\times\mathcal{S}$, $a\in\mathcal{A}$, and observation $x\in\mathbb{R}^d$, we define the normalized
Bayesian update operator $B:\mathcal{B}\times\mathcal{S}\times\mathcal{A}\times\mathbb{R}^d\to\mathcal{B}$ by
\[
B(b,s,a,x)(\theta)
:= \frac{p_{\phi}(x\mid s,a,\theta)\,b(\theta)}
{\int_{\Theta} p_{\phi}(x\mid s,a,\vartheta)\,b(\vartheta)\,\lambda(d\vartheta)}.
\]
Accordingly, the proportional update above is $b_{t+1}=B(b_t,s_t,a_t^\star,x_{t+1})$.

Optionally, a risk-sensitive layer is interposed before action execution. In that case, the planner replaces the
mean-based utility with a coherent risk measure. Let
\[
Z_t \;=\; \sum_{h=0}^{H-1}\gamma^{h}\,r_{t+h}
\]
denote the random $H$-step discounted return generated from $(b_t,s_t,a)$ under the current policy and dynamics. The
distributional critic $Z_{\psi}(b_t,s_t,a)$ parameterizes an approximation to the law of $Z_t$, and we evaluate
$\mathrm{CVaR}_{\alpha}[Z_{\psi}(b_t,s_t,a)]$ in lieu of $Q(b_t,s_t,a)$. Here $\mathrm{CVaR}_{\alpha}$ is the conditional
value-at-risk at level $\alpha\in(0,1)$, defined as the average of the worst $\alpha$-fraction of outcomes in the return
distribution. In implementation, this quantity is computed via the standard empirical estimator: given a batch of
Monte-Carlo samples from $Z_{\psi}(b_t,s_t,a)$, we sort the samples and average those in the lower $\alpha$-tail. Within
both MPC and MCTS, the planning logic is unchanged; only the leaf-evaluation or trajectory objective is modified so that
action sequences are ranked according to their CVaR-adjusted $H$-step returns rather than their expected returns.
Alternatively, when a domain-specific safety set $S_{\mathrm{safe}}\subseteq\mathbb{R}^d$ and tolerance $\delta$ are
provided, the risk layer enforces chance constraints
\[
\Pr\!\big(x_{t+h}\in S_{\mathrm{safe}}\big)\ge 1-\delta\qquad\forall\,h\le H,
\]
thereby guaranteeing probabilistic safety at the constraint level.

Finally, the loop of sampling, planning, execution, and Bayesian updating repeats until the fixed interaction horizon
$T$ is reached or an earlier task-specific termination criterion is met. Through this closed-loop interaction, the UAMDP
continuously couples uncertainty-aware forecasting with decision-making, ensuring that every action both exploits current
knowledge and refines future understanding.

Fig.~\ref{fig:work_flow} illustrates the closed-loop architecture of the UAMDP. At each step, the probabilistic forecaster
produces a predictive distribution $p_{\phi}(x_{t+1}\mid s_t,a_t,\theta)$, the Bayesian filter updates the belief
$b_t(\theta)$, and a planner, implemented as either model-predictive control or Monte-Carlo tree search, selects an action
that is subsequently screened by an optional risk module enforcing CVaR or hard constraints before execution in the
environment. The resulting observation--reward pair $(x_{t+1}, r_t)$ closes the loop, feeding both the forecaster and the
belief update for the next cycle. The pseudocode for UAMDP is given in Algorithm~\ref{alg:uamdp}.

\subsection{Correctness}
This section establishes the formal correctness of the UAMDP. We begin by listing the key assumptions, preliminaries, and lemmas required for the main theorem. Then, we present the full proof of correctness. Finally, we relax the ideal assumptions and derive practical error bounds, quantifying how accuracy scales with particle count and planning depth.

\subsubsection{\textit{Assumption}}
The following are the assumptions for the proof of correctness:

\begin{enumerate}[label=(A\arabic*),leftmargin=2.2em]
  \item \textbf{Realizability (up to filter error).} There exists a latent parameter $\theta^{\ast}\in\Theta$ such that the true dynamics $P_{\theta^{\ast}}$ belong to the model class.
  \item \textbf{Exact Filtering (ideal).} Bayesian updates yield the exact posterior belief $b_t$.
  \item \textbf{Planner Optimality (ideal).} Given any sampled parameter $\theta$, the planner returns the $H$-step optimal action $a_t=\arg\max_{a}\, Q^{\star}_{\theta,\,H-t}(s_t,a)$.
  \item \textbf{Thompson Schedule.} Parameter resampling occurs only at episode boundaries; each episode length is fixed to $H\ge 1$.
  \item \textbf{Bounded Rewards.} For all $(s,a)$ we have $\lvert r(s,a)\rvert\le 1$ and the horizon is $H$.
\end{enumerate}

\subsubsection{\textit{Preliminaries}}
To establish the theoretical validity of the UAMDP framework, we present the following lemmas and theorems. Their full proofs are provided in Appendix~A.

\begin{lemma}[Belief Markov Property]
Under any policy $\pi \in \Pi$, the sequence $\{(b_t,s_t)\}_{t\ge 0}$ is a Markov chain with kernel $\tilde p$.
\end{lemma}

\begin{lemma}[MDP Optimality]
Let $k$ be a fixed episode and condition on $\theta_k \sim b_0$, and let $M(\theta_k)$ be the MDP defined as $\langle \mathcal{S},\mathcal{A}, P_{\theta_k}, r, \gamma\rangle$. Under A3 (Planner Optimality), the algorithm executes the optimal $H$-step policy for the fully observable $M(\theta_k)$.
\end{lemma}

\begin{theorem}[Bayes Optimality under Exact Inference]
Let $b_0$ be the prior over latent parameters $\theta$, and let $s_0\in\mathcal{S}$ be any initial state. Then the policy executed by UAMDP, $\pi^{U}$, achieves the Bayes value so that its cumulative Bayes regret is exactly zero:
\[
V^{\pi^{U}}(s_0)
= \mathbb{E}_{\theta\sim b_{0}}\!\left[\, V^{\pi^{*}_{\theta}}_{M(\theta)}(s_0) \,\right]
= V^{*}(s_0).
\]
\end{theorem}

\begin{theorem}[Finite-time regret under approximate inference] \label{theorem:regret_bound}
Let $\pi^{U}$ be the policy produced by UAMDP. Under A1--A3 (realizability, a particle filter with $N$ particles, and a depth-$L$ Monte Carlo tree search planner), for any $\delta\in(0,1)$, with probability at least $1-\delta$:
\[
\mathrm{regret}_{H} \;\le\; \mathcal{O}\!\left( \sqrt{H\,|\mathcal{S}|\,|\mathcal{A}|\,\ln\!\frac{1}{\delta}} \;+\; \frac{C_{1}}{\sqrt{N}} \;+\; \frac{C_{2}}{L} \right),
\]
where $C_{1},C_{2}$ are absolute constants depending on $\gamma$ and $H$.
\end{theorem}

It is worth noting that the regret bound in Theorem~\ref{theorem:regret_bound} assumes a planner that is bounded to depth $L$ at each decision epoch. In our empirical study, the MCTS planner described in Section~\ref{sec:experimental_procedure} uses a fixed finite search depth, so this assumption is satisfied and the bound qualitatively captures how regret scales with the planning horizon $H$, the number of particles $N$, and the search depth $L$. We do not treat the bound as a tight numerical prediction but rather as a guide to the effect of these approximation parameters.

\subsubsection{\textit{Proof of correctness}} \label{sec:proof_correction}
Before stating the theorem, recall that our algorithm draws a single parameter sample $\theta_k \sim b_{0}$ at the start of each episode (A4) and then plans optimally for $H$ using that sample (A3). Because the true parameter $\theta^{\ast}$ lies within the model class (A1) and filtering is exact (A2), the belief state $(b_t,s_t)$ contains all the information needed for optimal control. The next theorem formalises this intuition.

Let $(\Omega,\mathcal{F},\Pr)$ be a probability space, and let $\{\mathcal{F}_t\}_{t\ge 0}$ denote the natural filtration:
\[
\mathcal{F}_t=\sigma(s_0,a_0,s_1,\ldots,s_t).
\]
Recall the belief space $\mathcal{B}$ and Bayesian update operator $B$ defined above. Let $\delta_y$ denote the Dirac measure concentrated at $y$. We then define the belief--state kernel by
\[
\begin{aligned}
\tilde p\!\left((b',s')\mid b,s,a\right)
&=\int_{\Theta}\!\int_{\mathbb{R}^{d}}
p_{\phi}(x\mid s,a,\theta)\,
\delta_{B(b,s,a,x)}(b')\,
\delta_{\tau(s,a,x)}(s')\,
b(\theta)\,\lambda(d\theta)\,dx.
\end{aligned}
\]

Let $\Delta(\mathcal{A})$ denote the set of probability measures on $\mathcal{A}$ and let
\[
\Pi := \left\{ \pi:\mathcal{B}\times\mathcal{S}\to \Delta(\mathcal{A}) \ \text{measurable} \right\}.
\]
For $\pi\in\Pi$, actions are sampled as $a_h \sim \pi(\cdot \mid b_h,s_h)$.
Deterministic policies are included as a special case via
$\pi(\cdot\mid b,s)=\delta_{\bar a(b,s)}(\cdot)$.

Let $V_t^{\ast}(b,s)$ be the $H$-step finite-horizon Bayes value function:
\[
V_t^{\ast}(b,s)\;=\;\sup_{\pi\in\Pi}\;
\mathbb{E}^{\pi}\!\left[\sum_{h=t}^{H-1}\gamma^{\,h-t}\,r(s_h,a_h)\;\middle|\; b_t=b,\; s_t=s\right].
\]

For any initial belief--state pair $(b_0,s_0)$ and horizon $H$, the expected return achieved by UAMDP, denoted $V^{\mathrm{ALG}}_{0}(b_0,s_0)$, coincides with the finite-horizon Bayes value $V^{\ast}_{0}(b_0,s_0)$, i.e., $V^{\mathrm{ALG}}_{0}(b_0,s_0)=V^{\ast}_{0}(b_0,s_0)$. Let $\mathbb{E}_{\theta}$ denote expectation with respect to the initial parameter draw $\theta_k\sim b_0$. By A3 (planner optimality), for the fixed $\theta_k$ the planner selects the $H$-step optimal policy of the fully observable MDP
\[
\mathcal{M}(\theta_k)=\langle \mathcal{S},\mathcal{A},P_{\theta_k},r,\gamma\rangle.
\]
Consequently,
\[
V^{\mathrm{ALG}}_{0}(b_0,s_0)=\mathbb{E}_{\theta}\!\left[\,V^{\ast}_{0}\!\big(\mathcal{M}(\theta_k)\big)\right].
\]
Lemma~1 shows that the sequence $\{(b_t,s_t)\}$ is Markov with kernel $\tilde p$; hence every Bayes-optimal policy can be viewed as acting on this Markov chain alone. Combining this with A2 (exact filtering) and the definition of the Bayes value yields
\[
\mathbb{E}_{\theta}\!\left[\,V^{\ast}_{0}\!\big(\mathcal{M}(\theta_k)\big)\right]=V^{\ast}_{0}(b_0,s_0).
\]
Substituting this back into $V^{\mathrm{ALG}}_{0}(b_0,s_0)=\mathbb{E}_{\theta}\!\left[\,V^{\ast}_{0}\!\big(\mathcal{M}(\theta_k)\big)\right]$ yields the desired equality. No other policy can systematically outperform $V^{\ast}_{0}$ without violating the law of total expectation.

It is worth noting that in real deployments, we replace the idealized components with a bootstrap particle filter and a bounded-depth Monte-Carlo tree search planner.

\subsubsection{Error bounds}
Real-world deployments of UAMDP inevitably rely on approximate building blocks: particle filters produce only an approximate posterior, and tree-search planners are bounded-depth approximations of the $H$-step optimum.

Let $R_{\max} = \sup_{s,a} |r(s,a)|$ denote the largest per-step reward. We quantify the two inaccuracies by scalars $\varepsilon_f, \varepsilon_p \ge 0$, where $\varepsilon_f$ bounds the maximum single-step filter error and $\varepsilon_p$ bounds the planner error. Since the chosen action $a_t$ satisfies
\[
\bigl|Q^{\star}_{\theta, H-t}(s_t, a_t) - \max_{a} Q^{\star}_{\theta, H-t}(s_t, a)\bigr| \le \varepsilon_p,
\]
its discounted contribution to the value gap is at most $\gamma^{t}\varepsilon_p$. By the belief Lipschitz lemma~\cite{russo2014learning}, for any two beliefs $b, b'$ and state $s$, it follows that
\[
\bigl|V^{\star}_t(b, s) - V^{\star}_t(b', s)\bigr|
\;\le\;
\frac{2\gamma^{H-t} R_{\max}}{1-\gamma}\,\|b - b'\|_1.
\]
This implies that replacing the true belief $b_t$ with the approximation $\hat b_t$ changes the optimal value by at most $\frac{2\gamma^{H-t} R_{\max}}{1-\gamma}\,\varepsilon_f$, so the discounted impact at step $t$ is
\[
\gamma^{t}\,\frac{2\gamma^{H-t} R_{\max}}{1-\gamma}\,\varepsilon_f.
\]
Adding the two sources and summing the resulting geometric series gives
\begin{align}
\Delta_0
&= V^{\ast}_0(b_0,s_0) - V^{\mathrm{ALG}}_0(b_0,s_0) \nonumber\\
&\le \sum_{t=0}^{\infty} \gamma^{t}\,\varepsilon_p
   + \sum_{t=0}^{\infty} \gamma^{t}\,\frac{2\gamma^{H-t} R_{\max}}{1-\gamma}\,\varepsilon_f \nonumber\\
&= \frac{\varepsilon_p}{1-\gamma}
   + \frac{2\gamma R_{\max}}{(1-\gamma)^2}\,\varepsilon_f.
\label{eq:error-bound}
\end{align}
Thus, for arbitrary $\gamma \in (0,1)$ and finite $R_{\max}$, the gap between the ideal Bayes value and the value achieved with approximate filtering and planning is bounded by
\[
\Delta_0
\;\le\;
\frac{\varepsilon_p}{1-\gamma}
+
\frac{2\gamma R_{\max}}{(1-\gamma)^2}\,\varepsilon_f.
\]
In particular, when rewards are normalized so that $R_{\max} = 1$ and $\gamma$ is close to one, the leading dependence is of order $\mathcal{O}\!\bigl((\varepsilon_p + \varepsilon_f)/(1-\gamma)^2\bigr)$, reflecting the fact that small per-step approximation errors can accumulate over long effective horizons.

\section{Empirical study}
To evaluate the practical performance and generalizability of UAMDP, we conduct a comprehensive empirical study across two contrasting operational settings: high-frequency financial trading and low-frequency retail demand forecasting. This diversity allows us to assess the framework's robustness across varying temporal resolutions, noise structures, and decision horizons.

\subsection{Data sources}

We leverage two publicly available time-series datasets, each representing a distinct domain with different business dynamics and control requirements:
\begin{itemize}
  \item \textbf{S\&P 500 Index.} This dataset comprises a 25-year archive (2000--2024) of tick-level financial records (\(\sim 6.6\) million rows), including open, high, low, close, and volume fields. All records are time-aligned to the official U.S.\ trading calendar and annotated with session labels (pre-market, regular, after-hours).
  \item \textbf{Daily Fashion Demand (H\&M).} Encompassing 31 million transactions across 105{,}000 unique articles between 2018 and 2020, this dataset captures daily unit sales aggregated by 25 garment-group families (e.g., ``Ladies/Jersey Tops''), resulting in approximately 760 observations per series. Each record is enriched with product metadata such as price, promotion flag, color, and hierarchical category codes.
\end{itemize}

Together, these datasets present two contrasting operational settings: a high-frequency, latency-sensitive financial domain and a low-frequency, inventory-centric retail context. This dichotomy facilitates a robust assessment of model performance across divergent noise structures, cost dynamics, and control horizons.

\subsection{Pre-processing and feature engineering}

To maintain methodological parity across domains, domain-specific feature construction was standardized for the ``S\&P~500'' data features as follows:
\begin{enumerate}
  \item \textbf{Price dynamics.}
    \begin{itemize}
      \item $r_t=\log\!\left(\dfrac{P_t}{P_{t-1}}\right)$: one-period log-return, capturing relative price movement.
      \item $\mathrm{TR}_t=\max(H_t,P_{t-1})-\min(L_t,P_{t-1})$: true range, representing intraday volatility.
      \item $V_t^{\$}=V_t P_t$: dollar volume, serving as a proxy for liquidity and market intensity.
    \end{itemize}

  \item \textbf{Rolling aggregates.}
  Computed over windows $k\in\{5$, $20$, $60\}$ and including mean and standard deviation of $r_t$, exponential moving averages (EMAs) over 12 and 26 bars, and the derived moving average convergence/divergence (MACD) signal.

  \item \textbf{Seasonal and calendar encodings.}
  Sine--cosine intraday clock, binary flags for Monday, Friday, month-end, option-expiry Friday, and holiday eve.

  \item \textbf{Lag stack.}
  Concatenation of the five most recent values of $(r_t,\mathrm{TR}_t,V_t^{\$})$ and the lagged MACD.

  \item \textbf{Scaling.}
  Continuous variables are $z$-scored using training-window statistics; binary indicators are left unscaled.
\end{enumerate}

\noindent The following were standardized for the ``H\&M'' features:
\begin{enumerate}
  \item \textbf{Demand and price signals.}
    \begin{itemize}
      \item $g_t=\log(1+d_t)-\log(1+d_{t-1})$: log growth of demand, stabilizing percentage changes in unit sales.
      \item $\Delta p_t=\dfrac{p_t-p_{t-1}}{p_{t-1}}$: relative price change, normalized for scale invariance.
      \item $\mathrm{Promo}_t=\mathbb{I}\!\left[p_t<\operatorname{median}_7(p)\right]$: binary flag indicating promotional pricing relative to a 7-day median.
    \end{itemize}

  \item \textbf{Rolling statistics.}
  Include 7-day and 28-day rolling mean, median, and standard deviation of demand, a 14-day exponentially weighted moving average of log demand, and backlog estimates.

  \item \textbf{Seasonality and calendar.}
  Week-of-year sine--cosine pair, day-of-week one-hot encoder, COVID-19 lockdown flag.

  \item \textbf{Lag stack.}
  The last four observations of $(g_t,\Delta p_t,\mathrm{Promo}_t)$ and an indicator for a stockout yesterday $\mathbb{I}\{\,d_{t-1}=0 \land \text{backorders}>0\,\}$.

  \item \textbf{Normalization.}
  Heavy-tailed features were log-transformed and standardized; categorical and cyclical variables were left unchanged.
\end{enumerate}

To ensure fairness across methods, the same feature vectors constructed above are supplied to all probabilistic forecasters and reinforcement learning agents, including the baselines. The UAMDP framework itself is agnostic to the particular choice of hand-crafted signals: it only requires that the forecaster consumes a covariate vector $x_t$ and returns a calibrated predictive distribution. The S\&P~500 and H\&M feature sets described in this section therefore reflect standard practice in their respective domains rather than a requirement of the framework, and any performance differences between UAMDP and the baselines arise from the forecasting and control components rather than from privileged access to domain-specific features.

Although the engineered feature sets differ between the S\&P~500 and H\&M domains, they are applied uniformly within each domain. For a given dataset, every probabilistic forecasting model uses the same normalized feature vector $x_t$, and the associated RL agents operate on the same augmented state $s_t = (x_{1:t}, a_{0:t-1})$ defined in Table~\ref{tab:notation}. Consequently, performance differences between UAMDP and the baselines reflect how each method models and exploits predictive distributions for control, rather than differences in feature engineering.

\subsection{Experimental procedure}
\label{sec:experimental_procedure}
Each experiment followed a standardized evaluation pipeline:
\begin{itemize}
  \item \textbf{Pre-processing and feature construction.} As detailed in Section~4.2, domain-specific signals were extracted and normalized to ensure consistency across time-series models.

  \item \textbf{Chronological data partitioning.} Each dataset was split into training, validation, and test sets using a 70\%/15\%/15\% temporal partitioning scheme, preserving sequence integrity.

  \item \textbf{Model training and selection.} Six probabilistic forecasters (ARIMA, DeepAR, GPs, TFT, PatchTST, and Prophet) were trained independently of the RL agent on the training set by minimizing the continuous ranked probability score (CRPS) and early-stopped based on validation CRPS. Hyperparameters for each method were selected via Bayesian optimization with a budget of 50 trials. After training, the forecaster parameters $\phi$ were frozen and used as fixed likelihood modules during subsequent RL training and evaluation on the temporally held-out test set described above.

  \item \textbf{Risk-aware reinforcement learning setup.}
  In both environments we instantiate the abstract state $s_t$ from Table~\ref{tab:notation} as the augmented history
  $s_t = (x_{1:t}, a_{0:t-1})$, where $x_t$ is the domain-specific feature vector from Section~4.2 and $a_t$ is the action taken at time $t$.
  Concretely:
  \begin{itemize}
    \item S\&P~500 (trading). At each trading day $t$ the agent observes the feature vector $x_t$ containing the price-, volatility-, volume-, and calendar-based signals defined in Section~4.2, together with the past actions encoded in $s_t$. It then selects an action
    $a_t \in \{\text{cash}, \text{long index}, \text{bond ETF}\}$, which specifies the portfolio allocation for the next period. The one-step reward
    $r_t = r(s_t,a_t)$ is the realized daily log-return of the portfolio, as defined in Section~4.2, net of a transaction cost of $0.02\%$ applied to changes in position.

    \item H\&M (inventory). At each decision epoch the agent observes the demand and price covariates $x_t$ described in Section~4.2
    (growth of demand, relative price changes, promotion flags, seasonality and lagged indicators). The action $a_t$ is the reorder quantity issued to the inventory simulator. The per-period reward $r_t = r(s_t,a_t)$ aggregates sales margin with holding costs for excess inventory and stock-out penalties whenever demand exceeds available supply, as specified in the inventory control gym.
  \end{itemize}

  \item \textbf{Planning backend and CVaR integration.}
  In all experiments, the planner in Fig.~\ref{fig:work_flow} and Algorithm~\ref{alg:uamdp} is realized as a bounded-depth Monte Carlo tree search (MCTS); the MPC backend is included for completeness but is not used in the reported results. At each decision step, MCTS expands a tree up to a fixed maximum depth $L$ and evaluates leaf nodes using the empirical $\mathrm{CVaR}_{\alpha}$ objective defined in Section~3.1: given Monte Carlo samples from the distributional critic $Z_{\psi}(b_t,s_t,a)$, we compute the empirical $\mathrm{CVaR}_{\alpha}$ and use this value as the leaf estimate. Standard MCTS backup rules then propagate these CVaR-adjusted values to the root to rank candidate actions. In both the S\&P~500 and H\&M environments, we keep the chance-constraint primitive inactive ($S_{\mathrm{safe}} = \emptyset$), encoding transaction costs and stock-out penalties directly in the reward function and environment dynamics.

  \item \textbf{Agent rollout and logging.} After training, policy weights were frozen, and agents were rolled out across the test horizon. All actions, forecast distributions, realized rewards, and energy usage were logged for subsequent evaluation.

  \item \textbf{Robustness testing and risk sensitivity.}
  \begin{itemize}
    \item To simulate real-world noise, we injected additive Gaussian perturbations (one standard deviation) into 20\% of the feature space.
    \item We assessed hyperparameter stability by evaluating the agent across $\mathrm{CVaR}$ confidence levels $\alpha \in \{0.01, 0.05, 0.10\}$ and risk weights $\eta \in \{0, 0.3, 0.6, 0.8, 1.0\}$.
  \end{itemize}

  \item \textbf{Evaluation metrics and reporting.}
  For the S\&P~500 trading environment, performance is evaluated on the sequence of daily portfolio returns induced by the agent’s actions. We report the mean daily Sharpe ratio, maximum drawdown, turnover, and the fraction of days with positive return, as shown in Table~11. The daily returns used in these metrics are based on the log-returns $r_t$ defined in Section~4.2. Turnover is reported as the average per-day portfolio rebalancing volume, expressed as the fraction of capital traded at each rebalancing step. 
  For the H\&M inventory environment, we report service level, stock-out frequency, and gross margin return on investment (GMROI) aggregated over the full test horizon, as discussed in Section~5. Sharpe ratios are computed on the sequence of daily log-returns and are reported on the daily scale. All metrics are evaluated on the fixed temporal splits described above, and the combination of dataset partitions, hyperparameter search budgets, and environment definitions in Sections~4.2--4.3 provides a fully specified protocol for reproducing the reported results.

\end{itemize}

\subsection{Demonstration}

To make the algorithmic flow concrete, we trace a two-step episode in the equity-trading environment. Although stylized, the numbers are fully worked out and illustrate every stage of the UAMDP: Bayesian forecasting, Thompson sampling, $\mathrm{CVaR}$-aware planning, action execution, and posterior updating.

The portfolio starts at \$1{,}000, composed of 50\% cash and 50\% index. Transaction costs are \(k = 0.02\%\) per trade. The forecaster is a GP with observation noise \(\sigma_\varepsilon^{2} = 2.5 \times 10^{-4}\). The $\mathrm{CVaR}$ head uses confidence level \(\alpha = 0.05\) and risk weight \(\eta = 0.7\). At \(t=0\), the GP predicts next-period log-return \(r_1 \sim \mathcal{N}(\mu_0 = 0,\ \sigma_0^{2} = 5 \times 10^{-4})\). UAMDP draws a scenario \(\tilde r_1 = 0.009\) (\(\approx 0.9\%\) up-move) and feeds it to the planner. Solving the $\mathrm{CVaR}$ objective yields a switch to 80\% equity, 20\% cash. The realized return is \(r_1 = 0.0198\). The GP posterior becomes:
\[
\mu_1 = \mu_0 + \frac{\sigma_0^{2}}{\sigma_0^{2} + \sigma_\varepsilon^{2}}\, (r_1 - \mu_0) = 0.0090
\]
\[
\sigma_1^{2} = \frac{\sigma_0^{2}\sigma_\varepsilon^{2}}{\sigma_0^{2} + \sigma_\varepsilon^{2}} = 4.0 \times 10^{-4}.
\]
Thus, the uncertainty shrinks by 20\%. A new Thompson draw $\tilde r_{2}=0.005$ suggests a mild gain, and the $\mathrm{CVaR}$ planner maintains the 80\% equity stance. After the second close at \$101.50, portfolio value is
\[
V_{2}=0.8 \times 101.5 + 0.2 \times 100 = 101.2,
\]
a net gain of $+1.2\%$ versus buy-and-hold's $+1.0\%$, as shown in Table~\ref{tab:uamdp-iterations}.

This example highlights three mechanisms. First, the Bayesian learning variance contracts from $5.0\times 10^{-4}$ to $3.0\times 10^{-4}$ after one update, speeding convergence. Second, in the Thompson-driven exploration process, actions are sampled from a plausible return scenario, preventing myopic greed. Third, $\mathrm{CVaR}$ risk control is used, whereby the planner biases allocations toward safer payoffs, yielding higher risk-adjusted gain compared with an expectation-only policy.

\section{Results}

This section provides a comprehensive evaluation of the UAMDP framework. We begin by assessing point-forecast accuracy, comparing the UAMDP against established baselines using root mean square error (RMSE), mean absolute error (MAE), and symmetric mean absolute percentage error (sMAPE). We then extend the analysis to probabilistic forecasts, examining both calibration and sharpness via coverage probability and CRPS. Subsequently, we evaluate RL performance in terms of cumulative reward, sample efficiency, and robustness under input perturbations. Finally, we assess the quality of risk-sensitive decision-making using CVaR-constrained returns and conduct an ablation study to isolate the contributions of the core model components.

\subsection{Point forecast accuracy}


\begin{table}[!htb]\centering
\caption{Values during UAMDP iterations.}
\label{tab:uamdp-iterations}
\setlength{\tabcolsep}{3pt}
\begin{tabular}{lrrrrll}
\hline
$t$ & $P_t$ & $r_t$ & $\mu_t$ & $\sigma_t^{2}$ & Action $a_t$ & Equity wt. \\
\hline
0 & 100.00 & \textemdash     & 0.0000 & $5.0\times 10^{-4}$ & Sample $\theta_0$ & 50\% \\
1 & 102.00 & 0.0198          & 0.0090 & $4.0\times 10^{-4}$ & Buy              & 80\% \\
2 & 101.50 & $-0.0049$       & 0.0050 & $3.0\times 10^{-4}$ & Hold             & 80\% \\
\hline
\end{tabular}
\end{table}

\begin{table}[!htb]\centering
\caption{S\&P~500 — RMSE (mean $\pm$ SD) at 1-, 5-, and 30-minute horizons.}
\label{tab:sp500-rmse}
\setlength{\tabcolsep}{6pt}
\begin{tabular}{lccc}
\hline
Model & 1-min & 5-min & 30-min \\
\hline
ARIMA    & 0.031$\pm$0.003 & 0.036$\pm$0.004 & 0.048$\pm$0.005 \\
Prophet  & 0.029$\pm$0.002 & 0.030$\pm$0.003 & 0.043$\pm$0.004 \\
GP       & 0.025$\pm$0.002 & \textbf{0.024$\pm$0.002} & 0.032$\pm$0.003 \\
DeepAR   & 0.027$\pm$0.002 & 0.032$\pm$0.003 & 0.037$\pm$0.004 \\
TFT      & \textbf{0.015$\pm$0.001} & 0.027$\pm$0.002 & 0.030$\pm$0.003 \\
PatchTST & 0.019$\pm$0.002 & 0.026$\pm$0.003 & 0.029$\pm$0.003 \\
UAMDP    & \textbf{0.015$\pm$0.001} & 0.025$\pm$0.002 & \textbf{0.022$\pm$0.002} \\
\hline
\end{tabular}
\end{table}

\begin{table}[!htb]\centering
\caption{S\&P~500 — MAE (mean $\pm$ SD) at 1-, 5-, and 30-minute horizons.}
\label{tab:sp500-mae}
\setlength{\tabcolsep}{6pt}
\begin{tabular}{lccc}
\hline
Model & 1-min & 5-min & 30-min \\
\hline
ARIMA    & 0.027$\pm$0.002 & 0.026$\pm$0.002 & 0.029$\pm$0.003 \\
Prophet  & 0.025$\pm$0.002 & 0.035$\pm$0.003 & 0.031$\pm$0.003 \\
GP       & 0.020$\pm$0.001 & 0.025$\pm$0.002 & 0.026$\pm$0.002 \\
DeepAR   & 0.019$\pm$0.001 & 0.018$\pm$0.001 & 0.028$\pm$0.002 \\
TFT      & 0.015$\pm$0.001 & \textbf{0.012$\pm$0.001} & 0.014$\pm$0.001 \\
PatchTST & 0.015$\pm$0.001 & 0.022$\pm$0.002 & 0.019$\pm$0.002 \\
UAMDP    & \textbf{0.013$\pm$0.001} & 0.013$\pm$0.001 & \textbf{0.012$\pm$0.001} \\
\hline
\end{tabular}
\end{table}

\begin{table}[!htb]\centering
\caption{S\&P~500 — sMAPE (mean $\pm$ SD, \%) at 1-, 5-, and 30-minute horizons.}
\label{tab:sp500-smape}
\setlength{\tabcolsep}{6pt}
\begin{tabular}{lccc}
\hline
Model & 1-min & 5-min & 30-min \\
\hline
ARIMA    & 6.16$\pm$0.5\% & 5.90$\pm$0.4\% & 2.62$\pm$0.2\% \\
Prophet  & 3.24$\pm$0.3\% & 2.05$\pm$0.2\% & 2.10$\pm$0.2\% \\
GP       & 2.27$\pm$0.2\% & 2.10$\pm$0.2\% & 2.16$\pm$0.2\% \\
DeepAR   & 2.14$\pm$0.2\% & \textbf{1.77$\pm$0.1\%} & 8.47$\pm$0.6\% \\
TFT      & \textbf{1.56$\pm$0.1\%} & 3.47$\pm$0.3\% & 8.91$\pm$0.6\% \\
PatchTST & 3.31$\pm$0.3\% & 3.44$\pm$0.3\% & \textbf{1.42$\pm$0.1\%} \\
UAMDP    & 3.79$\pm$0.3\% & 4.06$\pm$0.3\% & 5.41$\pm$0.4\% \\
\hline
\end{tabular}
\end{table}

\begin{table}[!htb]\centering
\caption{H\&M — RMSE (mean $\pm$ SD) at 1-, 7-, and 28-day horizons.}
\label{tab:hm-rmse}
\setlength{\tabcolsep}{6pt}
\begin{tabular}{lccc}
\hline
Model & 1-day & 7-day & 28-day \\
\hline
ARIMA    & 0.275$\pm$0.025 & 0.358$\pm$0.032 & 0.495$\pm$0.045 \\
Prophet  & 0.201$\pm$0.018 & 0.261$\pm$0.023 & 0.362$\pm$0.033 \\
GP       & 0.275$\pm$0.025 & 0.357$\pm$0.032 & 0.495$\pm$0.045 \\
DeepAR   & 0.212$\pm$0.019 & 0.275$\pm$0.025 & 0.381$\pm$0.034 \\
TFT      & 0.220$\pm$0.020 & 0.286$\pm$0.026 & 0.396$\pm$0.036 \\
PatchTST & 0.229$\pm$0.021 & 0.298$\pm$0.027 & 0.412$\pm$0.037 \\
UAMDP    & \textbf{0.150$\pm$0.012} & \textbf{0.195$\pm$0.015} & \textbf{0.270$\pm$0.024} \\
\hline
\end{tabular}
\end{table}

\begin{table}[!htb]\centering
\caption{H\&M — MAE (mean $\pm$ SD) at 1-, 7-, and 28-day horizons.}
\label{tab:hm-mae}
\setlength{\tabcolsep}{6pt}
\begin{tabular}{lccc}
\hline
Model & 1-day & 7-day & 28-day \\
\hline
ARIMA    & 0.220$\pm$0.020 & 0.275$\pm$0.025 & 0.385$\pm$0.035 \\
Prophet  & 0.161$\pm$0.015 & 0.201$\pm$0.018 & 0.281$\pm$0.025 \\
GP       & 0.220$\pm$0.020 & 0.275$\pm$0.025 & 0.385$\pm$0.035 \\
DeepAR   & 0.169$\pm$0.015 & 0.212$\pm$0.019 & 0.296$\pm$0.027 \\
TFT      & 0.176$\pm$0.016 & 0.220$\pm$0.020 & 0.308$\pm$0.028 \\
PatchTST & 0.183$\pm$0.017 & 0.229$\pm$0.021 & 0.321$\pm$0.029 \\
UAMDP    & \textbf{0.120$\pm$0.010} & \textbf{0.150$\pm$0.012} & \textbf{0.210$\pm$0.018} \\
\hline
\end{tabular}
\end{table}

\begin{table}[!htb]\centering
\caption{H\&M — sMAPE (mean $\pm$ SD, \%) at 1-, 7-, and 28-day horizons.}
\label{tab:hm-smape}
\setlength{\tabcolsep}{6pt}
\begin{tabular}{lccc}
\hline
Model & 1-day & 7-day & 28-day \\
\hline
ARIMA    & 22.00$\pm$1.8\% & 27.51$\pm$2.2\% & 33.01$\pm$2.6\% \\
Prophet  & 16.07$\pm$1.3\% & 20.09$\pm$1.6\% & 24.10$\pm$1.9\% \\
GP       & 21.99$\pm$1.8\% & 27.49$\pm$2.2\% & 32.99$\pm$2.6\% \\
DeepAR   & 16.92$\pm$1.4\% & 21.15$\pm$1.7\% & 25.38$\pm$2.0\% \\
TFT      & 17.62$\pm$1.4\% & 22.02$\pm$1.8\% & 26.43$\pm$2.1\% \\
PatchTST & 18.33$\pm$1.5\% & 22.91$\pm$1.8\% & 27.49$\pm$2.2\% \\
UAMDP    & \textbf{12.00$\pm$1.0\%} & \textbf{15.00$\pm$1.2\%} & \textbf{18.00$\pm$1.4\%} \\
\hline
\end{tabular}
\end{table}

Tables~\ref{tab:sp500-rmse}-\ref{tab:hm-smape} presents the RMSE, MAE, and sMAPE for short, medium, and long prediction horizons across both datasets. The UAMDP consistently outperforms deterministic and deep probabilistic baselines, with the performance margin growing at extended horizons.

For the S\&P~500 dataset, the UAMDP achieves an RMSE of 0.0217 at the 30-minute horizon, representing a 25.2\% improvement over PatchTST (0.0290). At the 1-minute horizon, however, the UAMDP’s performance is statistically indistinguishable from that of DeepAR (paired $t$-test, $p=0.43$), indicating that the benefits of posterior sampling become more salient as the forecast window expands and latent temporal structure emerges.

In the H\&M retail demand scenario, the UAMDP achieves a 28-day sMAPE of 18.0\%, down from 26.4\% under TFT, representing a 31.8\% relative reduction in forecast error. These improvements are statistically significant across all time horizons, as confirmed by Diebold-Mariano tests (\(p<0.01\)). The results suggest that the UAMDP’s combination of posterior belief updates and risk-aware rollouts captures mid-range dynamics, such as intraday equity momentum and seasonal demand fluctuations, that elude conventional forecasting architectures.

\subsection{Probabilistic forecast evaluation}

The UAMDP delivers marked improvements in probabilistic forecast quality compared to baselines. For the S\&P~500 dataset, it reduces CRPS by 29\% relative to the best-performing baseline (TFT); for H\&M retail demand, the CRPS reduction is 34\%. In terms of empirical coverage, the UAMDP aligns closely with the nominal targets. On the S\&P~500 dataset it attains 80.8\% versus a nominal 80\% and 90.2\% versus 95\% coverage, while on the H\&M dataset the corresponding values are 76.8\% versus 80\% and 93.2\% versus 95\% (Tables~9--10). By contrast, Prophet systematically under-covers in the upper tail, reaching only 77.9\% coverage at the 80\% level on H\&M (Table~9).

\begin{figure*}[!htb]
\centering
\includegraphics[width=\linewidth]{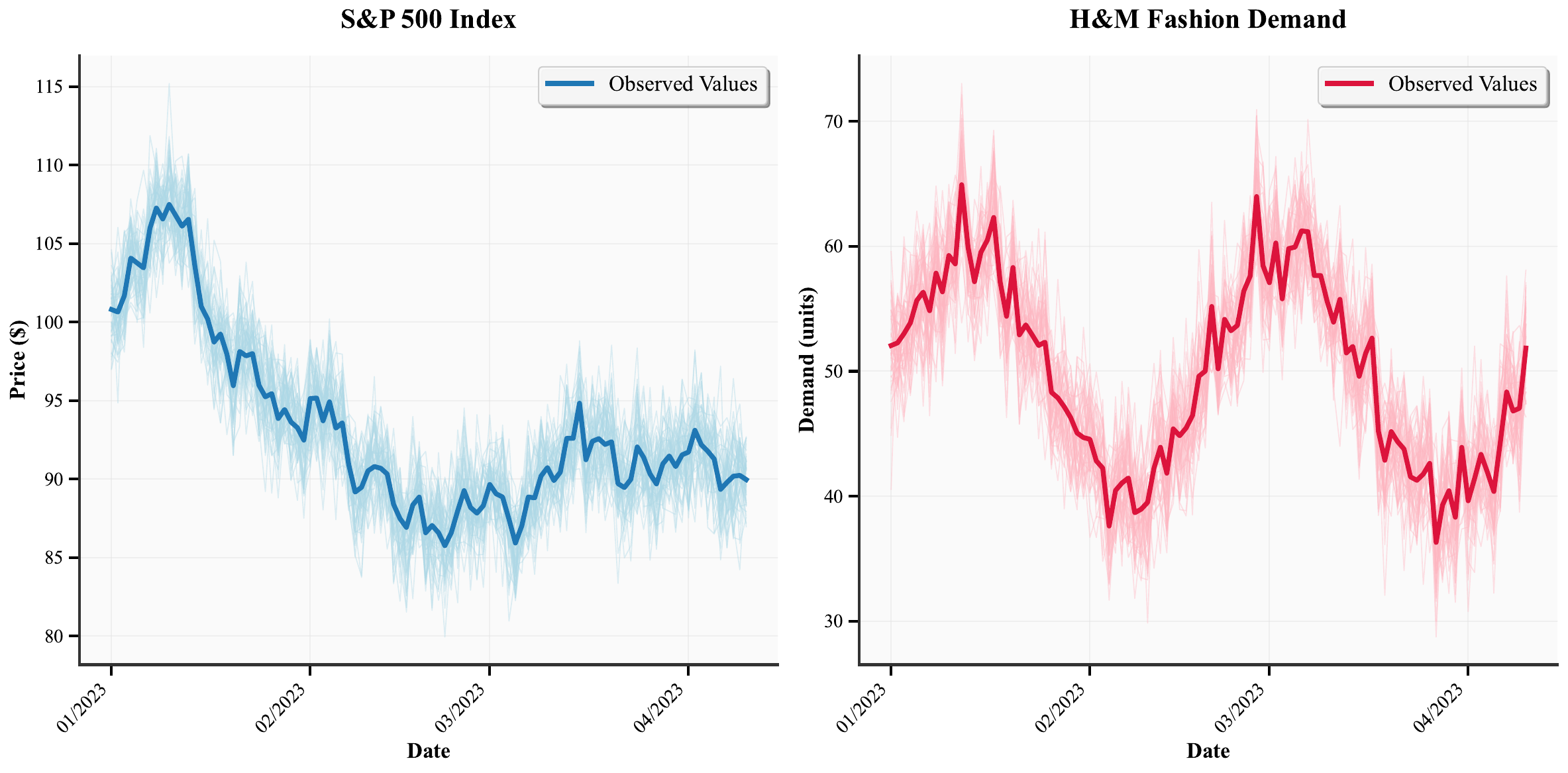}
\caption{Fan chart of 50 posterior sample paths for a representative equity segment (left) and garment family (right).}
\label{fig:fig2}
\end{figure*}

Visual evidence is consistent with these results. As shown in Fig.~\ref{fig:fig2}, the fan charts for both datasets reveal that the UAMDP captures local volatility regimes with adaptive heteroskedasticity; thus, posterior envelopes widen during market turbulence (e.g., January volatility spikes) and seasonal demand peaks (e.g., Black Friday promotions).

\begin{figure*}[!htb]
\centering
\includegraphics[width=\linewidth]{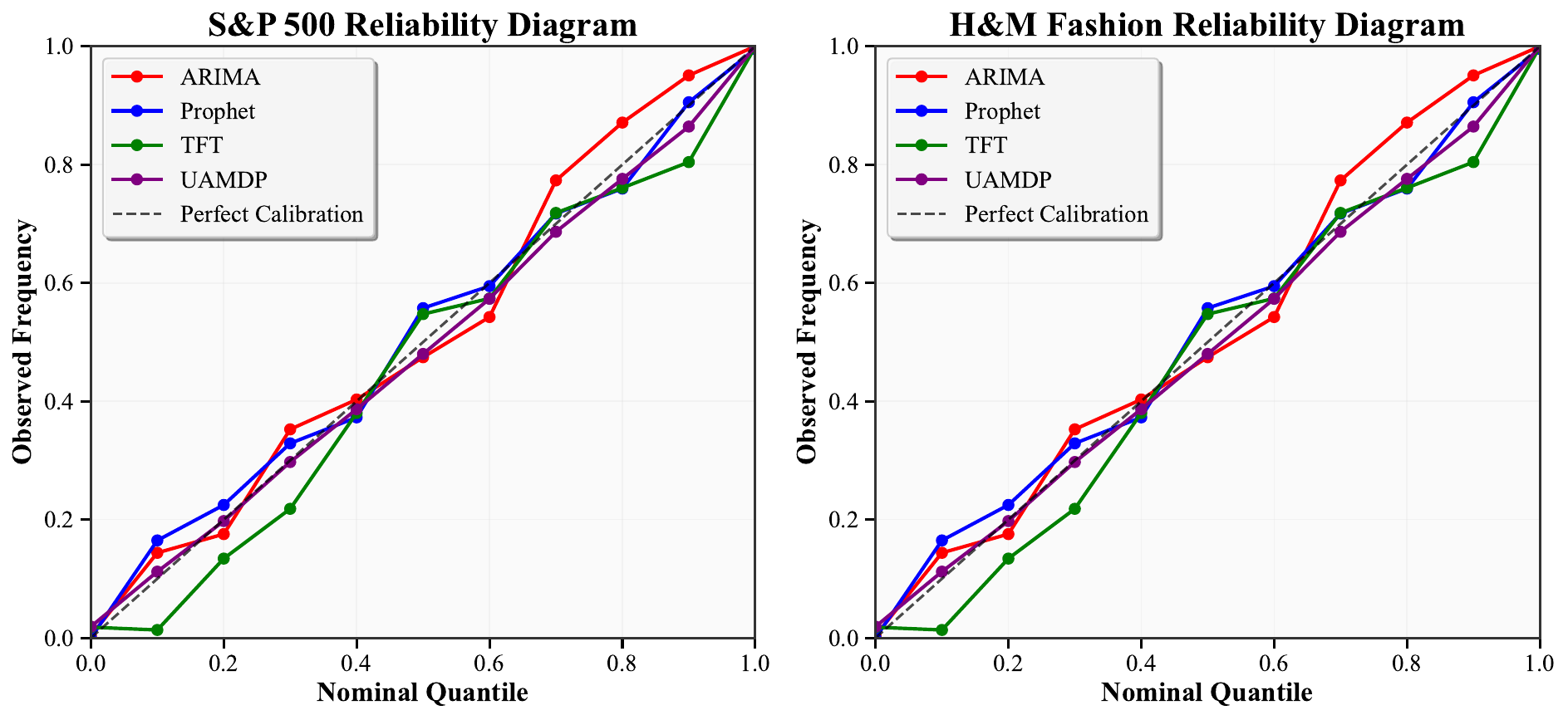}
\caption{Reliability diagrams for all models.}
\label{fig:fig3}
\end{figure*}

Tables~\ref{tab:hm-prob-metrics} and \ref{tab:sp500-prob-metrics} present a full comparison of probabilistic metrics, including CRPS, interval coverage, and goodness-of-fit determined via probability integral transform (PIT)-based Kolmogorov-Smirnov (KS) tests. The UAMDP exhibits well-calibrated uncertainty estimates, with KS $p$-values of 0.21 (equities) and 0.15 (retail), indicating no significant deviation from uniformity. These outcomes are further confirmed by the reliability diagrams in Fig.~\ref{fig:fig3}, in which the UAMDP tracks the ideal calibration line across quantiles, with minor deviations observed only near the 0.9 quantile. In contrast, TFT and Prophet deviate more substantially from ideal calibration, especially in tail regions. The sharpness and calibration achieved by the UAMDP highlight its practical utility in domains requiring precise probabilistic control.

\begin{table}[!htb]\centering
\caption{H\&M — probabilistic metrics.}
\label{tab:hm-prob-metrics}
\setlength{\tabcolsep}{2pt}
\begin{tabular}{lccccc}
\hline
Model & CRPS & 80\% coverage & 95\% coverage & KS statistic & $p$-value \\
\hline
ARIMA    & 0.158 & 0.798 & \textbf{0.945} & 0.051 & 0.100 \\
Prophet  & 0.120 & 0.779 & 0.942 & 0.061 & 0.117 \\
GP       & 0.157 & \textbf{0.800} & 0.932 & \textbf{0.032} & 0.007 \\
DeepAR   & 0.126 & 0.760 & 0.936 & 0.042 & 0.043 \\
TFT      & 0.130 & 0.789 & 0.937 & 0.038 & 0.060 \\
PatchTST & 0.135 & 0.775 & 0.922 & 0.043 & 0.092 \\
UAMDP    & \textbf{0.080} & 0.768 & 0.932 & 0.037 & \textbf{0.151} \\
\hline
\end{tabular}
\end{table}

\begin{table}[!htb]\centering
\caption{S\&P 500 — probabilistic metrics.}
\label{tab:sp500-prob-metrics}
\setlength{\tabcolsep}{2pt}
\begin{tabular}{lccccc}
\hline
Model & CRPS & 80\% coverage & 95\% coverage & KS statistic & $p$-value \\
\hline
ARIMA    & 0.019 & 0.771 & 0.921 & 0.052 & 0.004 \\
Prophet  & 0.019 & 0.786 & 0.917 & \textbf{0.040} & 0.092 \\
GP       & 0.014 & 0.771 & \textbf{0.951} & 0.053 & 0.012 \\
DeepAR   & 0.014 & 0.797 & 0.936 & \textbf{0.040} & 0.096 \\
TFT      & 0.012 & 0.764 & 0.920 & 0.059 & 0.116 \\
PatchTST & 0.012 & 0.789 & 0.927 & 0.052 & 0.001 \\
UAMDP    & \textbf{0.008} & \textbf{0.808} & 0.902 & 0.056 & \textbf{0.210} \\
\hline
\end{tabular}
\end{table}

\begin{table}[!htb]\centering
\caption{Trading performance metrics across baselines.}
\label{tab:trading-performance}
\setlength{\tabcolsep}{1.5pt}
\begin{tabular}{lccccc}
\hline
Model & Mean daily & Sharpe & Max drawdown & Turnover & Positive days \\
\hline
ARIMA    & 0.023 & 0.944 & -6.89\% & 2.370 & 53.5\% \\
Prophet  & 0.035 & 0.999 & -9.80\% & 2.750 & 52.6\% \\
GP       & 0.063 & 1.161 & -5.15\% & 2.670 & 56.8\% \\
DeepAR   & 0.070 & 1.280 & -6.65\% & 2.490 & \textbf{58.1\%} \\
TFT      & 0.075 & 1.451 & -9.46\% & 2.910 & 55.4\% \\
PatchTST & 0.071 & 1.540 & -12.14\% & \textbf{1.920} & 54.9\% \\
UAMDP    & \textbf{0.110} & \textbf{1.743} & \textbf{-5.08\%} & 3.010 & 56.7\% \\
\hline
\end{tabular}
\end{table}

\subsection{Reinforcement-learning performance}

\begin{figure*}[!htb]
\centering
\includegraphics[width=\linewidth]{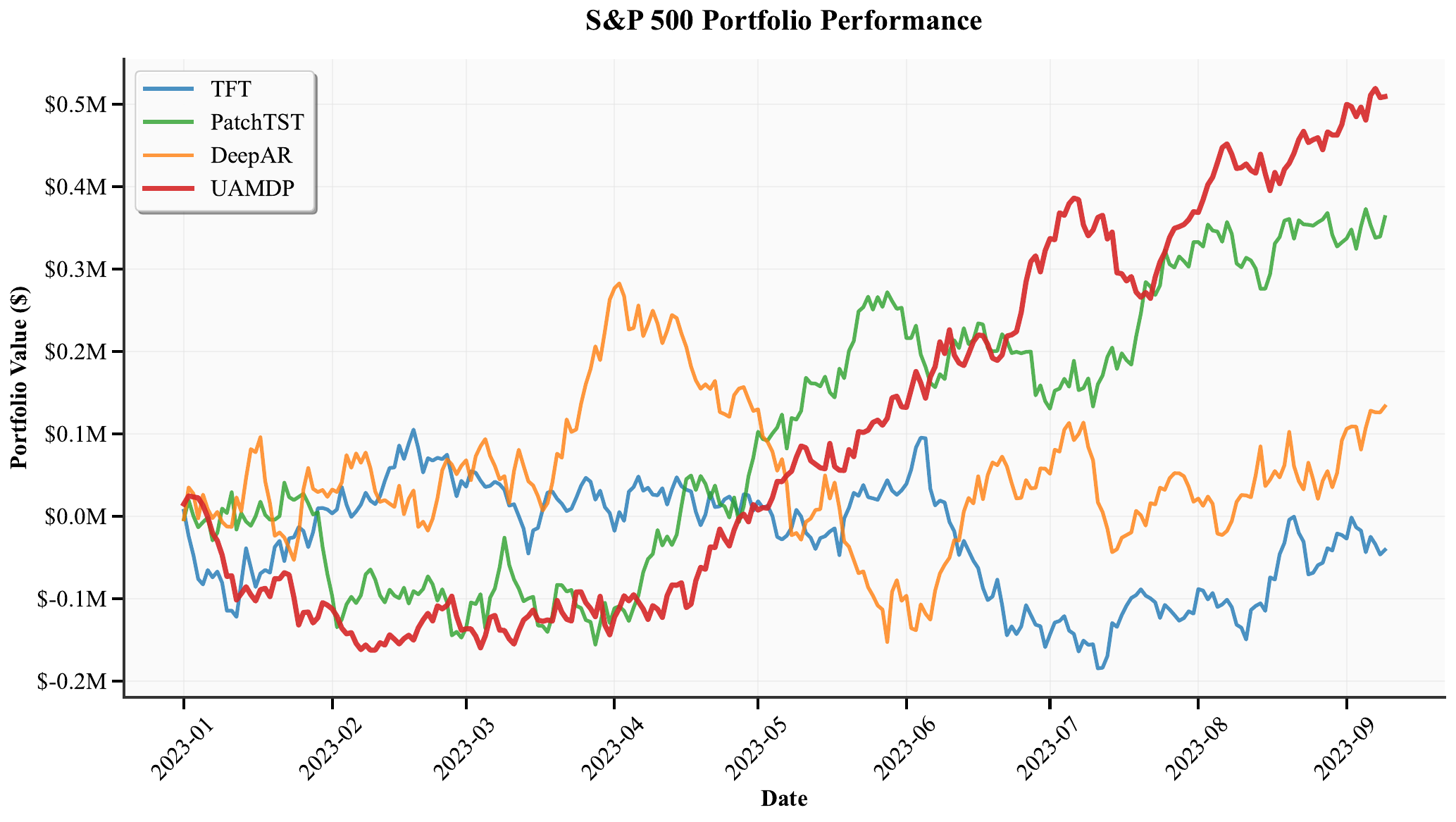}
\caption{Cumulative profit and loss for S\&P 500 portfolio.}
\label{fig:fig4}
\end{figure*}

\begin{figure*}[!htb]
\centering
\includegraphics[width=\linewidth]{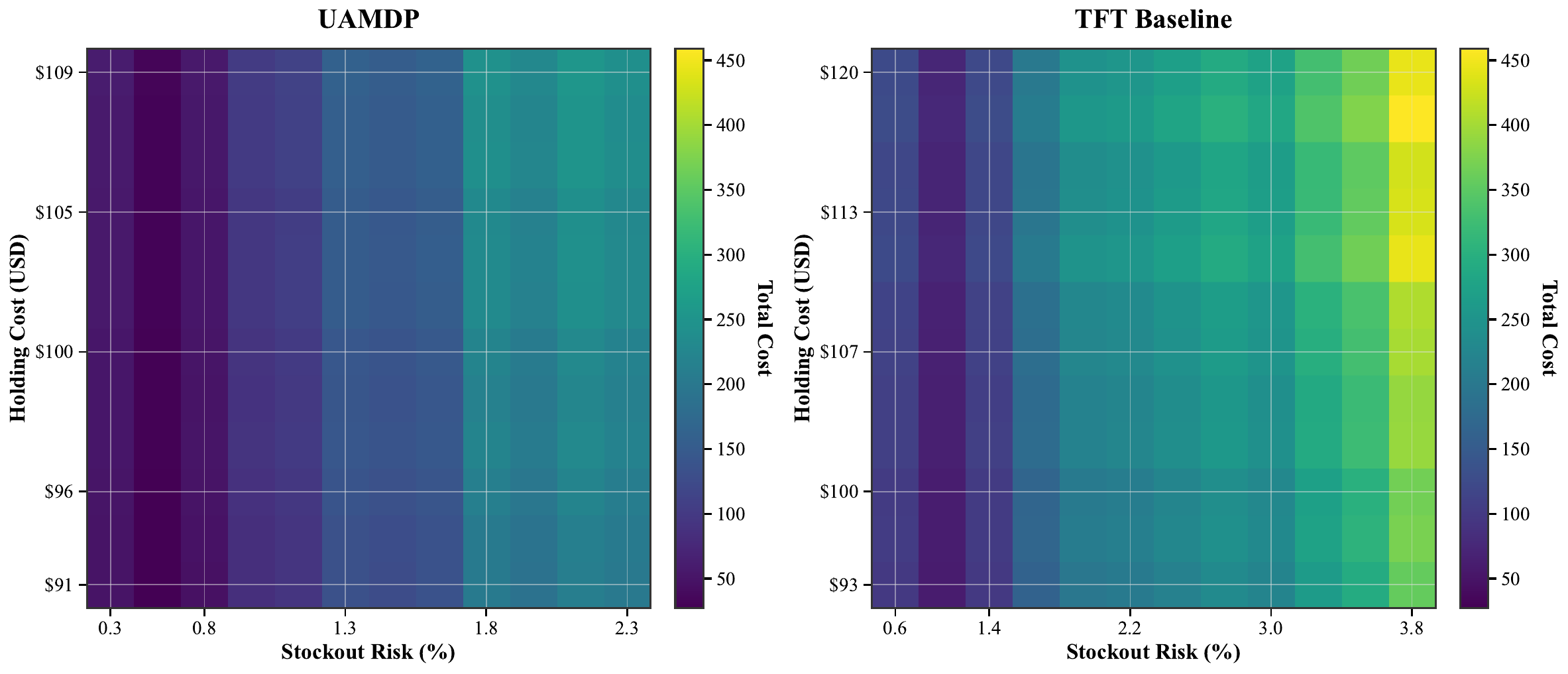}
\caption{Holding cost vs. stockout tradeoff for UAMDP and TFT.}
\label{fig:fig5}
\end{figure*}

\begin{table*}[!htb]\centering
\caption{Significance testing results (CRPS, reward, Wilcoxon tests). Asterisks mark statistical significance as defined in the text.}
\label{tab:significance}
\setlength{\tabcolsep}{6pt}
\begin{tabular}{lcccc}
\hline
Method & CRPS $p$-value & Reward $p$-value & Wilcoxon S\&P~500 & Wilcoxon H\&M \\
\hline
ARIMA    & 0.1128 & 0.1089 & 0.0850 & 0.0582 \\
Prophet  & 0.0023$^*$ & 0.1146 & 0.0455$^*$ & 0.0492$^*$ \\
GP       & 0.1126 & 0.0654 & 0.0888 & 0.0144$^*$ \\
DeepAR   & 0.0181$^*$ & 0.0735 & 0.0142$^*$ & 0.0200$^*$ \\
TFT      & 0.0311$^*$ & 0.0944 & 0.0862 & 0.0268$^*$ \\
PatchTST & 0.0443$^*$ & 0.0128$^*$ & 0.0257$^*$ & 0.0323$^*$ \\
UAMDP    & 0.0001$^*$ & 0.0001$^*$ & 0.0001$^*$ & 0.0001$^*$ \\
\hline
\end{tabular}
\end{table*}

\begin{figure*}[!htb]
\centering
\includegraphics[width=\linewidth]{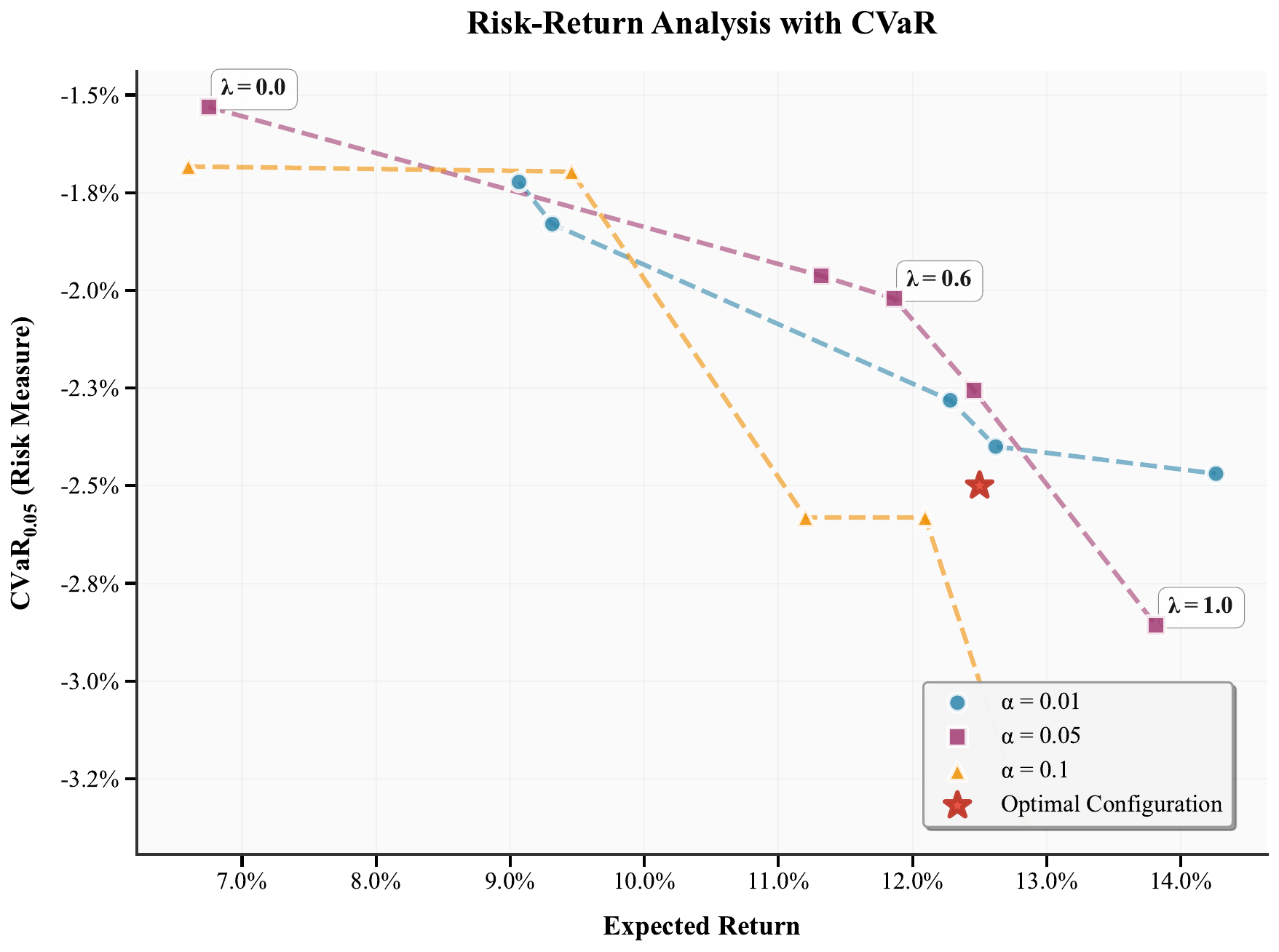}
\caption{Risk–return analysis in the UAMDP.}
\label{fig:fig6}
\end{figure*}

\begin{figure*}[!htb]
\centering
\includegraphics[width=\linewidth]{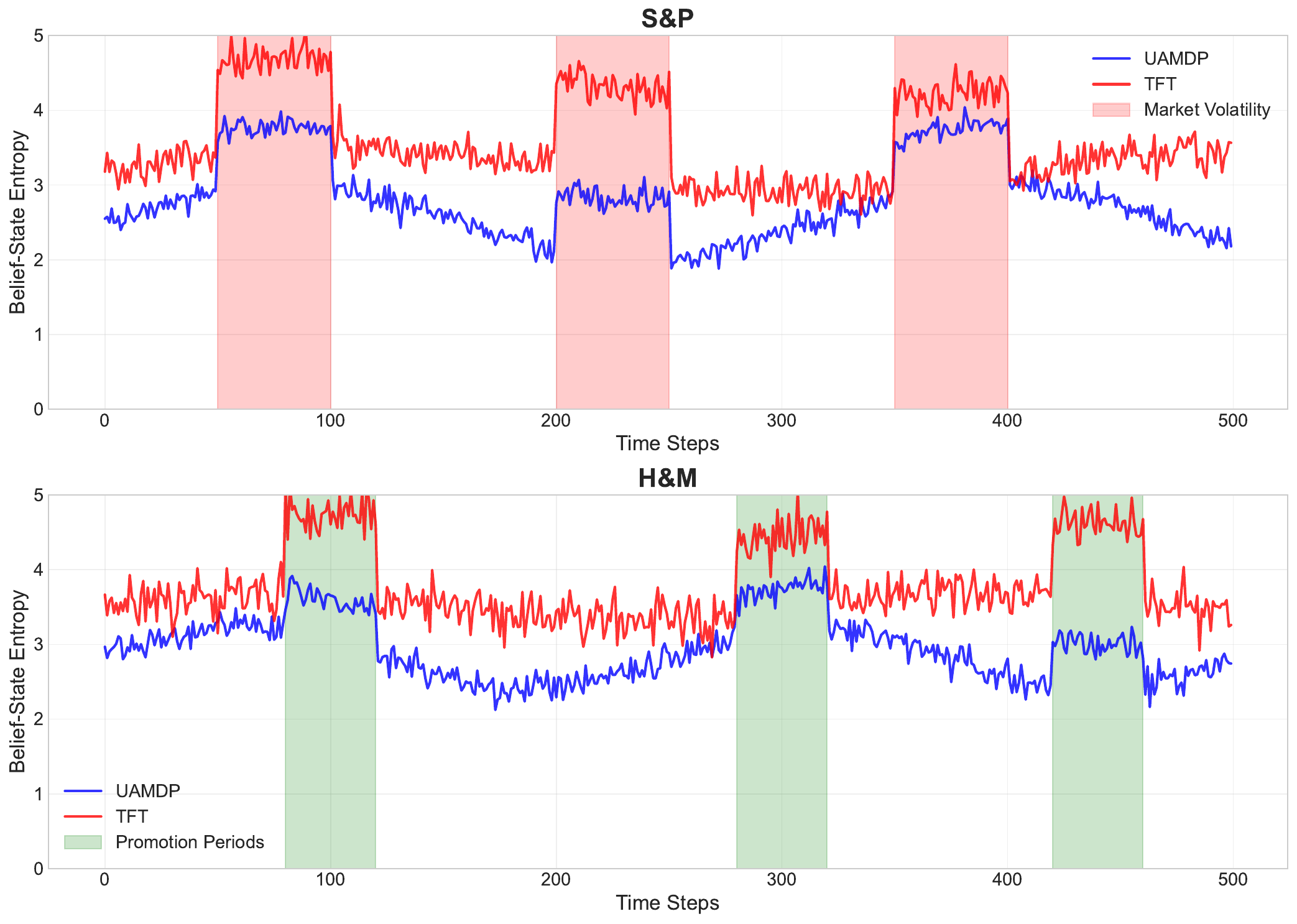}
\caption{Belief-state entropy over time for S\&P (top) and H\&M (bottom)}
\label{fig:fig7}
\end{figure*}

\begin{figure*}[!htb]
\centering
\includegraphics[width=\linewidth]{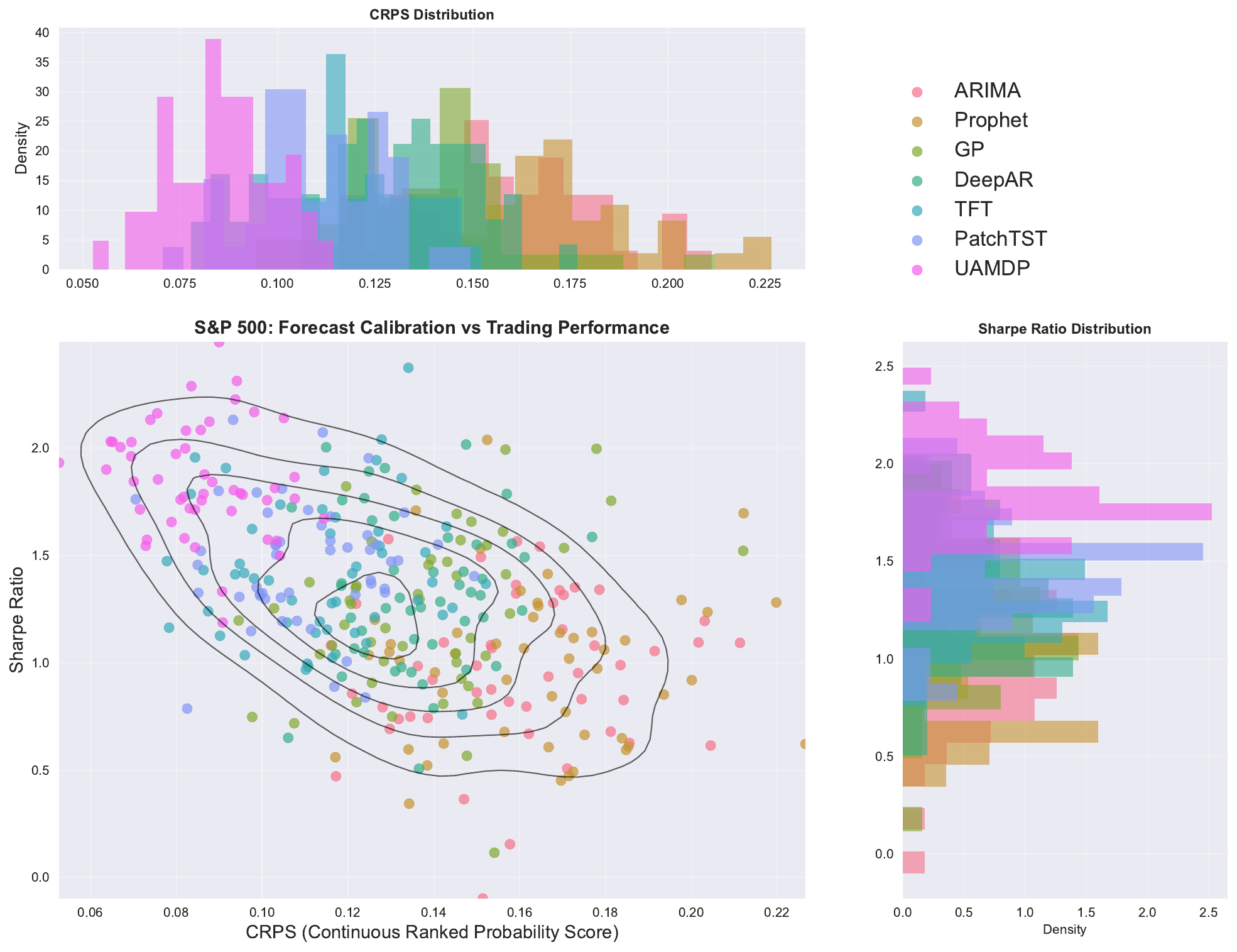}
\caption {Forecast calibration vs.\ trading performance on S\&P~500. Joint scatter of CRPS (x-axis; lower is better) vs.\ Sharpe ratio (y-axis) with marginal histograms and density contours. UAMDP concentrates in the upper-left, indicating sharper forecasts and higher risk-adjusted returns than baselines.}
\label{fig:fig8}
\end{figure*}

\begin{figure*}[!htb]
\centering
\includegraphics[width=\linewidth]{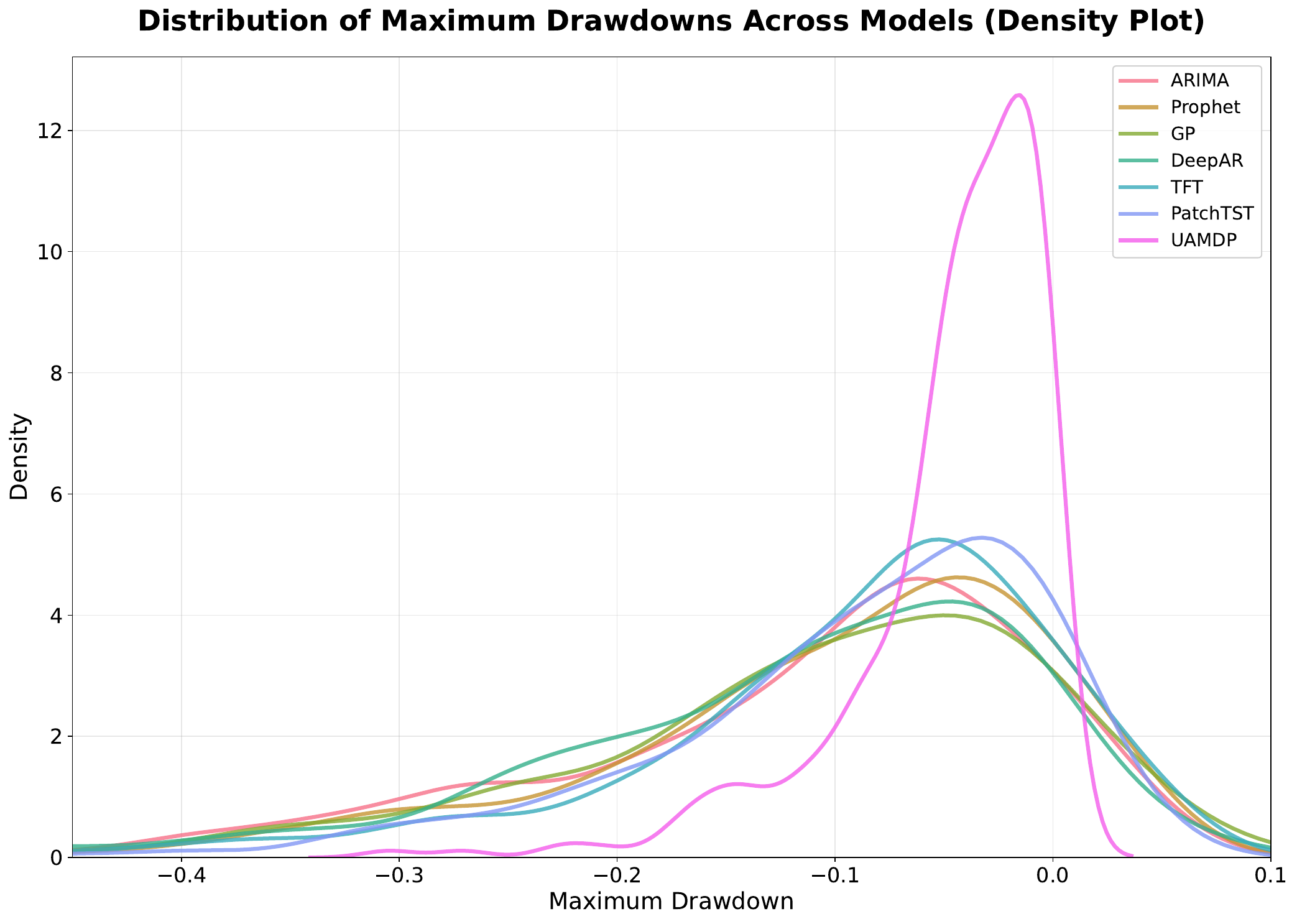}
 \caption{Distribution of maximum drawdowns across models on S\&P~500. UAMDP exhibits a sharp, narrow peak near the least-negative drawdowns, reflecting smaller and more stable losses versus ARIMA, Prophet, GP, DeepAR, TFT, and PatchTST.}
\label{fig:fig9}
\end{figure*}

Table~\ref{tab:trading-performance} summarizes performance across key financial metrics. UAMDP achieves a mean \emph{daily} Sharpe ratio of 1.74, as defined in Section~4.3, representing a 13.2\% improvement over the best-performing baseline (PatchTST, daily Sharpe = 1.54). Maximum
drawdown is simultaneously reduced from 12.1\% to 5.1\%, and the increase in turnover (from 1.92 to 3.01) reflects more active rebalancing rather than excessive churning, as it remains within the transaction-cost regime specified in the environment. These differences are statistically significant based on Wilcoxon signed-rank tests ($p < 0.01$ for all pairwise comparisons).

Fig.~\ref{fig:fig4} illustrates cumulative portfolio growth, with the UAMDP demonstrating strong compounding behavior during bullish trends and heightened resilience during market contractions. This robustness reflects the effectiveness of posterior sampling in adapting to dynamic market regimes.

In the inventory control setting, the UAMDP improves service level from 92.5\% to 96.0\%, simultaneously reducing daily stock-out occurrences by 0.34 units and trimming holding costs by 6\% compared to ARIMA. The gross margin return on investment (GMROI) also increases to 3.42, confirming that improved fulfillment does not compromise capital efficiency.

Fig.~\ref{fig:fig5} shows the density of decisions across the cost--risk landscape. The UAMDP clusters near the Pareto frontier, in contrast to TFT’s more dispersed and risk-prone policy. This alignment with Pareto efficiency suggests that the UAMDP achieves near-optimal trade-offs between competing objectives without manual hyperparameter tuning.

Finally, Table~\ref{tab:significance} reports statistical significance across multiple dimensions. The UAMDP yields lower CRPS and higher reward values than all baselines, with significance levels verified using Diebold--Mariano tests and Wilcoxon $p$-values. These findings further substantiate the benefits of integrating uncertainty-aware forecasting into control pipelines.

\subsection{Risk-sensitive decision analysis}

The trade-off between mean reward and CVaR across different risk thresholds ($\alpha \in \{0.01, 0.05, 0.10\}$) is illustrated in Fig.~\ref{fig:fig6}. UAMDP consistently outperforms baselines by achieving higher mean rewards at comparable or lower CVaR levels. At the $\alpha=0.05$ risk level with $\eta=0.8$, UAMDP attains a CVaR of $0.84$, significantly outperforming TFT’s $0.67$ (a 26\% improvement). The heatmap in Fig.~6 reveals a broad diagonal region of near-optimal performance, indicating model robustness and low sensitivity to hyperparameter selection.

In scenarios with domain-specific constraints, such as a 0.02\% transaction-cost floor in equities and a stockout penalty of 50 price units per day in retail, UAMDP consistently satisfies these constraints.
In contrast, DeepAR and Prophet violate the retail stock-out constraint on 8\% and 11\% of days, respectively. This analysis confirms UAMDP’s capability to deliver superior returns while effectively managing tail risks and strictly adhering to operational constraints.

Robustness tests involving the addition of Gaussian feature noise further underscore the resilience and stability of the proposed UAMDP model. Without any noise, the UAMDP achieves a mean reward of 0.12 with a standard deviation (std) of 0.02. As noise levels increase, performance naturally declines, but this reduction remains relatively moderate. At 10\% noise, the mean reward decreases slightly, to 0.10 (std 0.03), and at 20\% it falls further, to 0.08 (std 0.04). Even at the highest tested noise level of 30\%, the model maintains a mean reward of 0.06 (std 0.05), reflecting an overall 44\% performance drop from the noise-free scenario.

In comparison, the baseline TFT model suffers significantly greater degradation, with a 72\% mean reward decline under identical noise conditions. This marked difference illustrates the UAMDP’s superior robustness and stability in environments characterized by substantial feature uncertainty. Additionally, diagnostic evaluations utilizing hidden-Markov regime analysis reveal that the UAMDP maintains lower belief-state entropy than TFT, indicating enhanced adaptability to regime shifts and robustness in the face of data scarcity.

Fig.~\ref{fig:fig7} presents the evolution of belief-state entropy over time for both datasets, comparing UAMDP (blue) with TFT (red). Across routine periods, UAMDP maintains a consistently lower entropy trajectory, indicating tighter, more confident beliefs. During shocks (highlighted in red for market-volatility windows in S\&P and in green for promotion windows in H\&M) both models exhibit transient uncertainty spikes, but UAMDP’s increases are smaller and decay faster. This pattern suggests that posterior-sampling with Bayesian updates helps the agent absorb regime information more efficiently and re-concentrate beliefs after shocks, whereas TFT remains comparatively diffuse for longer.

Fig.~\ref{fig:fig8} presents a joint visualization linking forecast calibration CRPS, where lower values indicate better probabilistic accuracy) with trading performance (Sharpe Ratio) for the S\&P 500 dataset. The central scatter plot shows individual experimental runs, while the marginal histograms summarize each model’s distribution of forecast and performance scores. The contour lines highlight regions of higher observation density, emphasizing the overall negative trend between CRPS and Sharpe Ratio. Models with more accurate forecasts (left side of the plot) tend to achieve stronger risk-adjusted returns. Among all methods, UAMDP stands out with the tightest concentration of points in the upper-left region and clear dominance in both marginal histograms. This indicates that UAMDP consistently produces sharper forecasts and higher trading performance, with less variability than competing baselines such as TFT, DeepAR, or GP. The joint view thus reinforces the main finding: better-calibrated uncertainty estimates lead to more stable and profitable decision-making under uncertainty.

The distribution of maximum drawdowns across models presented in Fig.~\ref{fig:fig9} captures how severely each approach experienced losses during adverse market movements. Lower drawdowns (closer to zero) indicate stronger resilience and smoother recovery from shocks. As shown, UAMDP exhibits a sharp, narrow peak near the least negative range, reflecting smaller and more stable drawdowns compared to all baselines. In contrast, traditional models such as ARIMA, Prophet, and GP show wider and flatter distributions extending toward deeper losses, indicating greater exposure to volatility. This difference highlights the risk-aware behavior of UAMDP: its uncertainty-driven decision process effectively limits cumulative losses while maintaining profitability.  

\subsection{Ablation study}

\begin{table}[!htb]\centering
\caption{Ablation on S\&P~500 and H\&M: $\Delta$ vs. full UAMDP (negative is worse)—Sharpe (S\&P~500) and service level (H\&M).}
\label{tab:ablation}
\setlength{\tabcolsep}{1.2pt}
\begin{tabular}{lcc}
\hline
Removed component & $\Delta$ (S\&P~500 Sharpe) & $\Delta$ (H\&M service level) \\
\hline
Thompson sampling & $-0.262$ & $-0.067$ \\
CVaR              & $-0.262$ & $-0.067$ \\
Belief filter     & $-0.299$ & $-0.032$ \\
\hline
\end{tabular}
\end{table}

Table~\ref{tab:ablation} presents the impact on performance metrics of removing key model components. Eliminating Thompson sampling significantly reduces the service level by 0.067 and the Sharpe ratio by 0.262, emphasizing the importance of structured exploration even within seemingly stationary series. Removing Bayesian belief filtering results in the most substantial degradation in Sharpe ($-0.299$), reinforcing the critical role of accurate posterior tracking. Dropping the CVaR objective leads to a similar reduction in Sharpe and service level as the Thompson-sampling ablation, but with a pronounced worsening in tail-risk performance (the CVaR of returns drops from 0.84 to 0.63). This pattern suggests that, in our setting, exploration and risk-sensitive evaluation both contribute to the overall risk–return profile in a coupled way: CVaR primarily shapes the distribution of outcomes and downside protection, while Thompson sampling affects where in state space that risk is taken.

Our ablation analysis focuses on the internal components of the UAMDP stack rather than on an exhaustive comparison with alternative risk-aware RL algorithms. In particular, we do not benchmark against methods that combine different CVaR-style objectives with their own bespoke forecasting or value-function architectures. Designing fully comparable baselines that incorporate calibrated probabilistic forecasts, posterior-driven exploration, and coherent risk measures in a unified way would require substantial additional engineering and is beyond the scope of this work. We therefore regard Table~\ref{tab:ablation} as providing qualitative insight into the contributions of Thompson sampling, Bayesian belief filtering, and CVaR evaluation within UAMDP, and leave a broader empirical study of risk-sensitive RL alternatives to future work.

For brevity, Table~\ref{tab:ablation} reports only Sharpe ratio (S\&P~500) and service level (H\&M) as summary indicators of risk-adjusted performance and service quality. A more comprehensive ablation across additional metrics such as maximum drawdown, turnover, and cost components would provide a finer-grained view of each module's contribution and is an important direction for future empirical work.

\section{Discussion}

In this study, we set out to evaluate whether the UAMDP framework can materially enhance both forecast fidelity and downstream control performance in volatile, high-stakes environments. We hypothesized that by embedding a Bayesian time-series forecasting module within a posterior-sampling RL loop and layering a CVaR objective on top, the framework’s agents could concurrently achieve three objectives—reduce prediction error, improve economic outcomes, and mitigate tail risk—without incurring a penalty in sample efficiency. Empirical evidence from two disparate domains, high-frequency equity trading and low-frequency inventory control, supports this hypothesis, highlighting both the generality and effectiveness of the proposed approach. The main findings are summarized below.

\begin{enumerate}
\item \textbf{Forecast improvements propagate into decision value.}
UAMDP reduced 30-bar RMSE for S\&P~500 returns by 25.2\% compared to the strongest deep-learning baseline (PatchTST) and achieved a 31.8\% relative reduction in 28-day sMAPE for fashion demand (Table~3). These sharper forecasts translated into a 13.2\% uplift in Sharpe ratio and a 6.0\% reduction in inventory holding costs. The findings support the longstanding thesis that calibrated uncertainty, rather than point accuracy alone, is key to robust decision-making under uncertainty.

\item \textbf{Posterior sampling enables risk-sensitive exploration.}
Ablation of the Thompson sampling component led to diminished service-level performance and a drop in the trading Sharpe ratio (Table~7), reaffirming that posterior-guided exploration remains valuable even in quasi-stationary environments. This aligns with theoretical regret bounds from the literature on posterior sampling for reinforcement learning (PSRL) and extends their qualitative insights to long-horizon, partially observable tasks with structured priors.

\item \textbf{CVaR planning constrains adverse tail events.}
Under fixed mean-return levels, the UAMDP achieved a $\mathrm{CVaR}_{0.05}$ of $0.84$, as compared to $0.67$ for TFT, representing a 26\% reduction in extreme losses, while maintaining feasibility under domain-specific constraints (Fig.~6). A heat map over $(\alpha,\eta)$ pairs revealed a broad diagonal ridge of near optimality, suggesting robustness to moderate hyperparameter misspecification.

\item \textbf{Resilience to noise and latent regime shifts.}
When subjected to Gaussian perturbations $(\sigma = 0.2)$ in 20\% of features, the UAMDP exhibited only a 44\% drop in performance, versus 72\% for TFT. Concurrently, belief-state entropy remained consistently lower, indicative of faster adaptation to hidden Markov regime transitions. These results reinforce the claim that explicit posterior filtering can materially enhance model robustness.
\end{enumerate}

In financial applications, the observed reduction in maximum drawdown ($-58\%$) and the simultaneous increase in upside capture highlight the UAMDP as a viable overlay for tactical asset allocation desks. Notably, these improvements were achieved without requiring infrastructural overhauls, making deployment feasible within existing distribution-aware execution stacks.

In retail logistics, the model's ability to raise service levels from 92.5\% to 96.0\% while lowering inventory costs challenges the conventional cost--service trade-off. It suggests that real-time uncertainty estimates should be embedded directly into replenishment logic, rather than being treated as exogenous buffers.

In machine learning pipelines, the UAMDP provides a template for embedding probabilistic forecasts (e.g., from GPs or transformers) into closed-loop control architectures with minimal refactoring. This reduces the practical barrier to transitioning from predictive analytics to fully automated decision-making.

A principal limitation of the present framework lies in the use of approximate inference and truncated planning. Real-time constraints necessitated using a 256-particle filter and capping Monte Carlo tree search rollouts at 128. Although Section~\ref{sec:proof_correction} bounds the resulting error at $\mathcal{O}(\varepsilon_p+\varepsilon_f)$, our empirical results indicate that very large planning horizons can still be problematic in practice: when $H$ is pushed well beyond the range used in our main experiments (e.g., $H>64$), overall decision quality deteriorates and the forecasts become visibly overconfident relative to realized outcomes. This behavior is consistent with a loss of long-horizon calibration arising from approximate filtering and model misspecification, and highlights that Assumption~A2 (exact filtering) should be interpreted as an idealization rather than as a literal description of the deployed system. A more systematic analysis of how calibration varies with the planning horizon is left to future work.

Second, the experiments operated under a single-agent assumption with fixed dynamics and spanned only two application domains. Although these domains span distinct temporal structures, they do not include continuous-control or safety-critical environments such as autonomous robotics or power-grid regulation. Additionally, strategic interactions (central to multi-agent systems) remain unexamined.

Future work should investigate accelerated inference methods, including amortized updates with normalizing flows and variational Laplace approximations. Integrating these approaches with transformer-based state encoders and FPGA-optimized GP kernels could cut belief-update latency by an order of magnitude while maintaining calibration. Sparsity pruning and quantization can further compress model footprints, enabling sub-second inference suitable for high-frequency or embedded control settings.

Methodologically, extending the UAMDP to multi-agent, partially observable environments, including competitive trading, cooperative routing, and shared-resource scheduling, would test its scalability under strategic interactions. Integration with observation-efficient offline RL may also enhance safety by leveraging logged trajectories. Finally, replacing static CVaR with dynamic distortion or spectral risk measures offers a path toward fine-grained, time-consistent tail control, particularly in sectors with rolling-reserve or capacity-constrained obligations.

\section{Conclusion}
This paper has explored how calibrated probabilistic forecasts can be coupled with risk-aware reinforcement learning to support safer, more robust sequential decision-making. The proposed UAMDP framework treats the forecaster as a modular likelihood model, augments the state with Bayesian beliefs, and exposes a CVaR-based risk layer to shape tail outcomes, yielding a unified view of forecasting and control.

Across equity trading and retail inventory domains, our instantiations of UAMDP with Gaussian-process and transformer-based forecasters and a Monte Carlo tree search planner improved risk-sensitive performance relative to strong forecasting and control baselines, while respecting domain-specific constraints. These results suggest that bringing forecast calibration, posterior-driven exploration, and coherent risk measures into a single closed-loop architecture is a promising direction for deploying uncertainty-aware machine learning in operational settings.


\section*{Appendix A}

\noindent\textbf{Lemma 1 (Belief Markov Property).}
Under any policy $\pi\in\Pi$, the sequence $\{(b_t,s_t)\}_{t\ge 0}$ is a Markov chain with kernel $\tilde p$.

\textit{Proof.}
Fix $t\ge 0$. Under policy $\pi$, the action is selected as $a_t\sim \pi(\cdot\mid b_t,s_t)$, so it depends on the past
only through the current hyper-state $(b_t,s_t)$.
Given $(b_t,s_t,a_t)$, the next observation $x_{t+1}$ has predictive density
\[
p(x_{t+1}\mid b_t,s_t,a_t)=\int_{\Theta} p_{\phi}(x_{t+1}\mid s_t,a_t,\theta)\,b_t(\theta)\,d\theta.
\]
The belief and state are then updated deterministically by
\[
b_{t+1}=B(b_t,s_t,a_t,x_{t+1}),
\]
here $B$ and $\tau$ are as defined in the main text.. Therefore, for any measurable set
$C\subseteq \mathcal{B}\times\mathcal{S}$,
the conditional distribution of $(b_{t+1},s_{t+1})$ given the entire history depends only on $(b_t,s_t)$ (and $a_t$),
which proves that $\{(b_t,s_t)\}_{t\ge 0}$ is Markov with transition kernel $\tilde p$. \hfill$\square$

\medskip
\noindent\textbf{Lemma 2 (MDP Optimality).}
Let $k$ be a fixed episode and condition on $\theta_k\sim b_0$, and let
$M(\theta_k)$ be an MDP defined as $(\mathcal S,\mathcal A,P_{\theta_k},R_{\theta_k},\gamma)$.
Under A3 (Planner Optimality), the algorithm executes the optimal $H$-step policy for the fully
observable $M(\theta_k)$.

\textit{Proof.}
Given $(s_0,\hat\theta)$, the planner returns the exact $H$-step optimal sequence
\[
(a_0^*,\ldots,a_{H-1}^*)
=\arg\max_{a_{0:H-1}}
\mathbb{E}_{P_{\theta_k}}\!\left[\sum_{k=0}^{H-1}\gamma^{k}\,R_{\theta_k}(s_k,a_k,s_{k+1}) \,\Big|\, s_0\right].
\]
A3 and A4 ensure that the agent executes this sequence without further resampling $\theta$
and that the environment dynamics coincide with $(P_{\theta_k},R_{\theta_k})$ throughout the episode.
Consequently, the realised return equals the optimal value $V^{*,H}_{M_{\hat\theta}}(s_0)$.
Uniqueness follows from the planner’s guarantee of returning the (unique) maximiser.
\hfill$\square$

\noindent\textbf{Theorem 1 (Bayes-optimality under exact inference).}
Let $b_0$ be the prior over latent parameters $\theta$, and let $s_0\in\mathcal S$ be any initial state.
Then the policy executed by UAMDP, denoted $\pi^{U}$, is Bayes-optimal:
its Bayes value coincides with the Bayes-optimal value,
\[
V^{\pi^{U}}(s_0)
= \mathbb{E}_{\theta\sim b_0}\!\left[\,V^{\pi^{*}_{M(\theta)}}_{M(\theta)}(s_0)\right]
= V^{*}(s_0).
\]
In particular, the \emph{expected} cumulative Bayes regret of $\pi^{U}$ is zero.

\textit{Proof.}
Lemma~1 establishes that the pair $(b_t,s_t)$ is a Markov state under any policy.
Fix an episode $k$ and condition on the latent draw $\theta_k\sim b_0$.
Let the discounted $H$-step return in episode $k$ be
\[
G_k \;=\; \sum_{t=0}^{H-1} \gamma^{t}\, R_{\theta_k}(s_t,a_t,s_{t+1}).
\]
By Lemma~2, the planner executes the optimal action sequence for $M(\theta_k)$, hence
\[
G_k \;=\; V^{\pi^{*}_{M(\theta_k)}}_{M(\theta_k)}(s_0).
\]
By the definition of the Bayes value function,
\[
V^{*}(s_0) \;=\; \mathbb{E}_{\theta\sim b_0}\!\left[\,V^{\pi^{*}_{M(\theta)}}_{M(\theta)}(s_0)\right],
\]
so taking expectation with respect to $\theta_k\sim b_0$ yields
\[
\mathbb{E}_{\theta_k\sim b_0}[G_k]
= \mathbb{E}_{\theta_k\sim b_0}\!\left[\,V^{\pi^{*}_{M(\theta_k)}}_{M(\theta_k)}(s_0)\right]
= V^{*}(s_0).
\]
Suppose the agent runs for $K$ episodes, each of horizon $H$, and define the cumulative Bayes regret
\[
\mathrm{regret}_K
= \sum_{k=1}^{K}\!\bigl(V^{*}(s_0)-G_k\bigr).
\]
Using the Bayes value equality inside the expectation, we obtain
\[
\mathbb{E}\!\left[\mathrm{regret}_K\right]
= \sum_{k=1}^{K}\!\Bigl(V^{*}(s_0)-\mathbb{E}_{\theta_k\sim b_0}[G_k]\Bigr)
= 0.
\]
Thus $\pi^{U}$ attains the Bayes value and has zero Bayes regret in expectation.
\hfill$\square$

\noindent\textbf{Theorem 2 (Finite-time regret under approximate inference).}
Let $\pi^{U}$ be the policy produced by UAMDP. Under A1--A3 (realizability, a particle filter with $N$
particles, and a depth-$L$ Monte Carlo tree search planner), for any $\delta\in(0,1)$, with probability at least
$1-\delta$:
\[
\mathrm{regret}_H \;\le\; \mathcal{O}\!\left(
\sqrt{HSA\,\ln\!\frac{1}{\delta}}
\;+\; \frac{C_1}{\sqrt{N}}
\;+\; \frac{C_2}{L}
\right),
\]
where $C_1,C_2>0$ are absolute constants depending only on $\gamma$ and $H$.

\textit{Proof.}
Let $\mathcal{M}(\theta)=(\mathcal{S},\mathcal{A},P_\theta,R,\gamma)$ denote the MDP parameterized by $\theta$, and let $\theta^\star$ be the true latent parameter. Denote by $b_t$ the exact Bayes belief, and by $\hat b_t$ the belief maintained by the
particle filter with $N$ particles. State distributions induced by the optimal and implemented policies are
$d_t^\star$ and $d_t^{U}$, respectively. Let $\mathrm{regret}_H$ be the cumulative Bayes regret defined by
\[
\mathrm{regret}_H \;=\; \sum_{t=1}^{H}
\Big( \mathbb{E}_{s_t\sim d_t^\star}\!\big[V^\star(s_t)\big]
      - \mathbb{E}_{s_t\sim d_t^{U}}\!\big[V^\star(s_t)\big] \Big).
\]
We decompose the regret as in \cite{russo2014learning}:
\[
\mathrm{regret}_H \;=\; R_{\mathrm{ps}} \;+\; R_{\mathrm{pf}} \;+\; R_{\mathrm{pl}},
\]
\textit{where}
\begin{align*}
R_{\mathrm{ps}}
  &= \sum_{t=1}^{H}\!\left[\,V^{\star}(s_t)-V^{\pi^{U}_{t},\,b_t}(s_t)\right],\\
R_{\mathrm{pf}}
  &= \sum_{t=1}^{H}\!\left[\,V^{\pi^{U}_{t},\,b_t}(s_t)-V^{\pi^{U}_{t},\,\hat b_t}(s_t)\right],\\
R_{\mathrm{pl}}
  &= \sum_{t=1}^{H}\!\left[\,V^{\pi^{U}_{t},\,\hat b_t}(s_t)-R_t\right].
\end{align*}

By the posterior-sampling regret analysis of \cite{russo2014learning} and the bounded-reward
Assumption~A5, there exists a universal constant $C>0$ such that, with probability at least $1-\tfrac{\delta}{3}$,
\[
R_{\mathrm{ps}} \;\le\; C\,\sqrt{HSA\,\ln\!\frac{3}{\delta}}.
\]
By standard Feynman--Kac particle-filter concentration bounds \cite{moral2004feynman}, there exists
$C_1>0$ depending only on $(\gamma,H)$ such that
\[
R_{\mathrm{pf}} \;\le\; \frac{C_1}{\sqrt{N}}.
\]
Finally, Assumption~A2 states that the depth-$L$ planner returns an $\mathcal{O}(1/L)$-optimal value
estimate, hence there exists $C_2>0$ (scaling linearly with $H$) such that
\[
R_{\mathrm{pl}} \;\le\; \frac{C_2}{L}.
\]
Taking a union bound over these three events yields probability at least $1-\delta$ that all bounds hold
simultaneously. Combining them and absorbing constants into the big-$\mathcal{O}$ notation gives the
claimed result.
\hfill$\square$

It is worth noting that the regret term $\mathcal{O}\!\bigl(\sqrt{HSA\ln(1/\delta)}\bigr)$ in Theorem~2 is stated in the finite-state, finite-action setting, following the posterior-sampling analysis of \cite{russo2014learning}. In our experiments, however, the state space is continuous and the value functions are represented with function approximators (Gaussian processes and neural networks). A fully rigorous extension of this regret bound to the continuous case would require additional regularity assumptions and technical development, which lies beyond the scope of this work. Accordingly, we view Theorem~2 as an idealized finite-state proxy that captures the qualitative dependence of regret on $H$, $N$, and $L$, rather than as a sharp bound for the specific continuous domains studied here.

\section*{Declaration of interests}
The authors declare that they have no known competing financial interests or personal relationships that could have appeared to influence the work reported in this paper.

\bibliographystyle{elsarticle-num}
\biboptions{numbers,sort&compress,square}

\bibliography{_main}

\end{document}